\documentclass[10pt,twocolumn,letterpaper]{article}

\usepackage[pagenumbers]{cvpr}

\usepackage[dvipsnames]{xcolor}

\definecolor{cvprblue}{rgb}{0.21,0.49,0.74}
\usepackage[pagebackref,breaklinks,colorlinks,citecolor=cvprblue]{hyperref}
\usepackage{multirow}
\usepackage{overpic}
\usepackage{amsmath}
\usepackage{bm}
\usepackage{makecell}
\usepackage{tikz}
\usepackage{tabularx}
\usepackage{graphicx}
\usepackage[export]{adjustbox}
\usepackage{wrapfig}
\usepackage{amsmath}
\usepackage{mathrsfs}

\newcommand{\tb}[1]{\textbf{#1}}

\newcommand{\mpage}[2]
{
\begin{minipage}{#1\linewidth}\centering
#2
\end{minipage}
}

\title{$L_0$-Sampler: An $L_{0}$ Model Guided Volume Sampling for NeRF}

\author{\large Liangchen Li \quad Juyong Zhang\thanks{Corresponding author} \vspace{0.5 mm}\\
{\normalsize University of Science and Technology of China}\\
{\tt\small \{vvmno180662@mail., juyong@\}ustc.edu.cn} \\
}

\begin{document}

\twocolumn[{
\renewcommand\twocolumn[1][]{#1}
\maketitle
\begin{center}
\vspace{-30pt}
    \centering
    \captionsetup{type=figure}

    % \begin{subfigure}{0.59\textwidth}
    % \centering
    % \includegraphics[width=1.\linewidth]{figures/results/teaser/teaser_1.pdf}
    % \label{teaser_1}
    % \end{subfigure}
    % \hspace{.04in}
    % \begin{subfigure}{0.39\textwidth}
    % \centering
    % \includegraphics[width=1.\linewidth]{figures/results/teaser/teaser_2.pdf}
    % \label{fig:teaser_2}
    % \end{subfigure}
    
    \includegraphics[width=.59\linewidth]{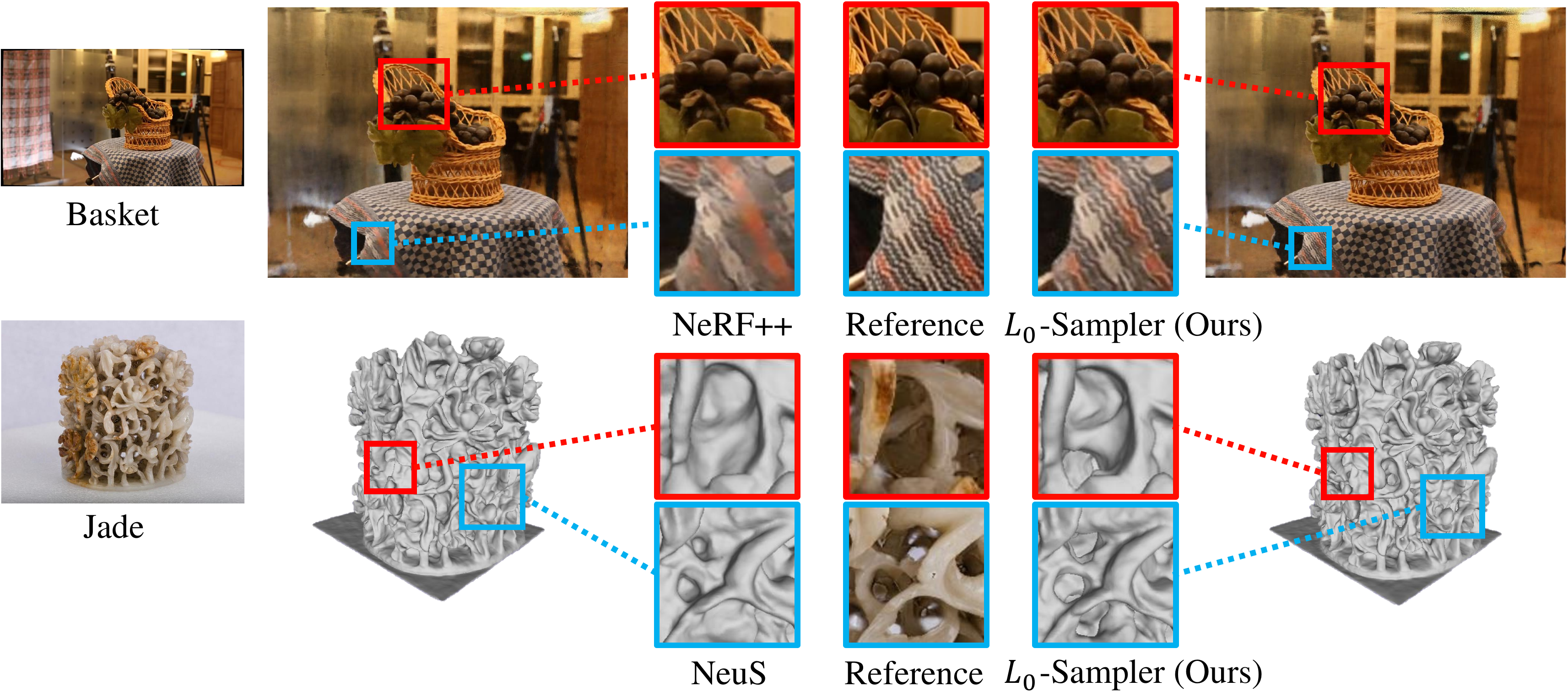}
    \hspace{.04in}
    \includegraphics[width=.39\linewidth]{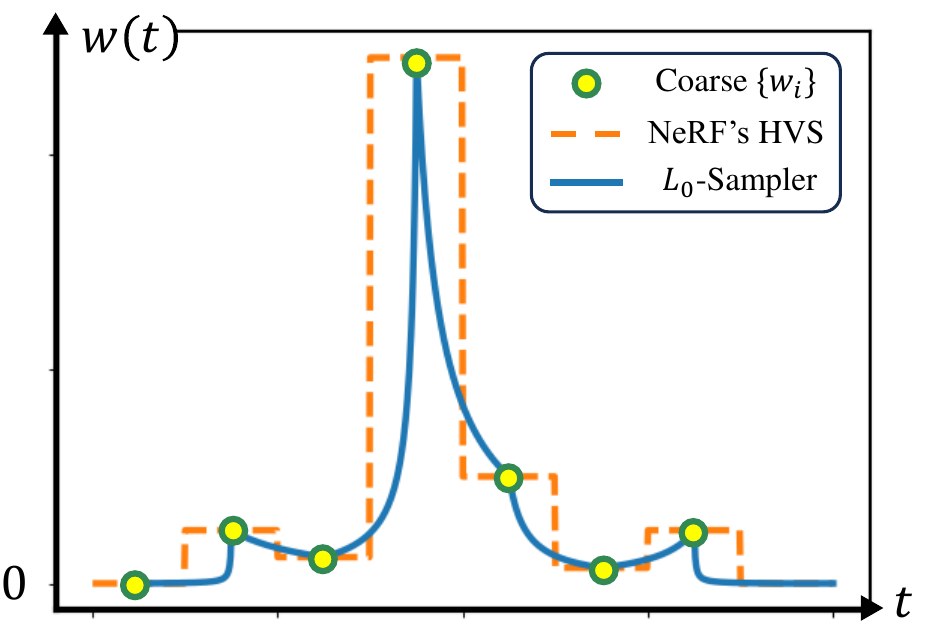}
    \captionof{figure}{We present $L_0$-Sampler, which is an upgrade of the Hierarchical Volume Sampling strategy of NeRF. By testing on different data sets, our proposed $L_0$-Sampler with different NeRF frameworks can achieve stable performance improvements on rendering and reconstruction tasks with few lines of code modifications and around the same training time. Left: Result comparison between HVS and our $L_0$-Sampler. Right: Instead of using piecewise constant functions when fitting $w(t)$, we use piecewise exponential functions for interpolation to get a quasi- $L_0$ $w(t)$, resulting in more concentrated and precise sampling.}
    \label{fig:teaser}
    \end{center}
}]

{
  \renewcommand{\thefootnote}%
    {\fnsymbol{footnote}}
  \footnotetext[1]{Corresponding Author}
}
\begin{abstract}

Since being proposed, Neural Radiance Fields (NeRF) have achieved great success in related tasks, mainly adopting the hierarchical volume sampling (HVS) strategy for volume rendering. However, the HVS of NeRF approximates distributions using piecewise constant functions, which provides a relatively rough estimation. Based on the observation that a well-trained weight function $w(t)$ and the $L_0$ distance between points and the surface have very high similarity, we propose $L_0$-Sampler by incorporating the $L_0$ model into $w(t)$ to guide the sampling process. Specifically, we propose to use piecewise exponential functions rather than piecewise constant functions for interpolation, which can not only approximate quasi-$L_0$ weight distributions along rays quite well but also can be easily implemented with few lines of code without additional computational burden. Stable performance improvements can be achieved by applying $L_0$-Sampler to NeRF and its related tasks like 3D reconstruction. Code is available at \url{https://ustc3dv.github.io/L0-Sampler/}.

\end{abstract}    
\section{Introduction}
\label{sec:intro}

\begin{figure}[htbp]
\centering
\begin{subfigure}{0.43\textwidth}
    \centering
    \includegraphics[width=\linewidth]{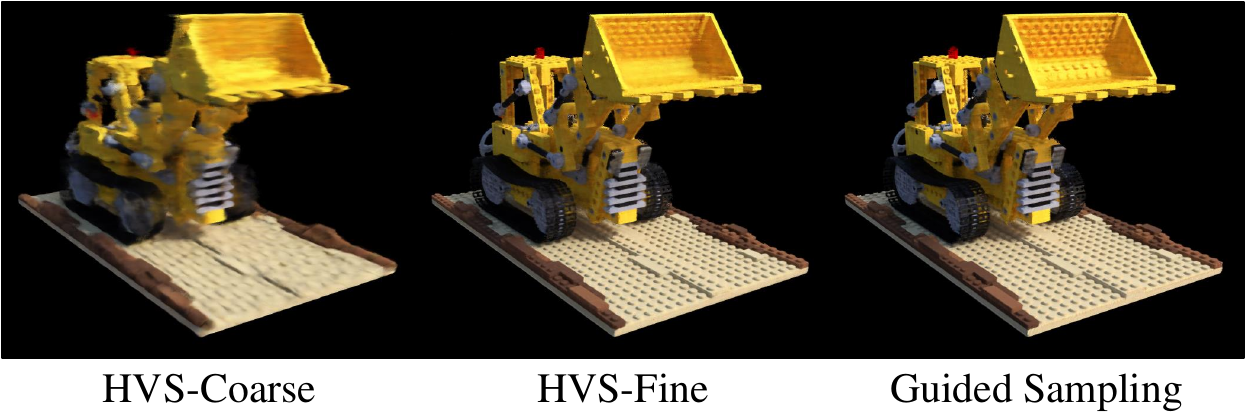}
    \caption{Visual Comparison}
    \label{fig:guide}
  \end{subfigure}
\begin{subfigure}{0.23\textwidth}
    \centering
    \includegraphics[width=\linewidth]{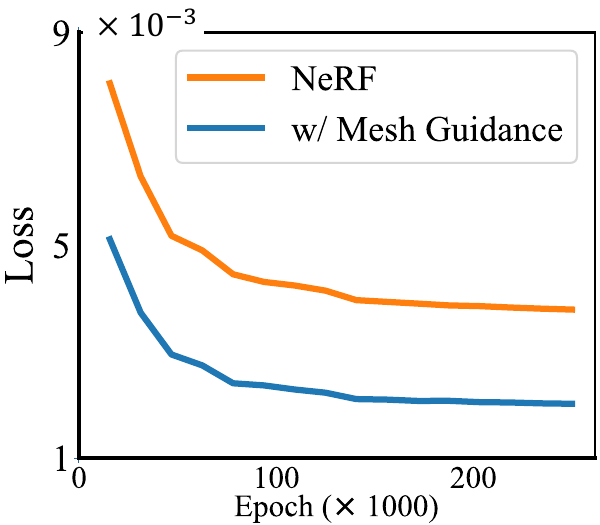}
    \caption{Rendering Loss}
    \label{fig:losses}
  \end{subfigure}
  % \hfill
  \begin{subfigure}{0.23\textwidth}
    \centering
    \includegraphics[width=\linewidth]{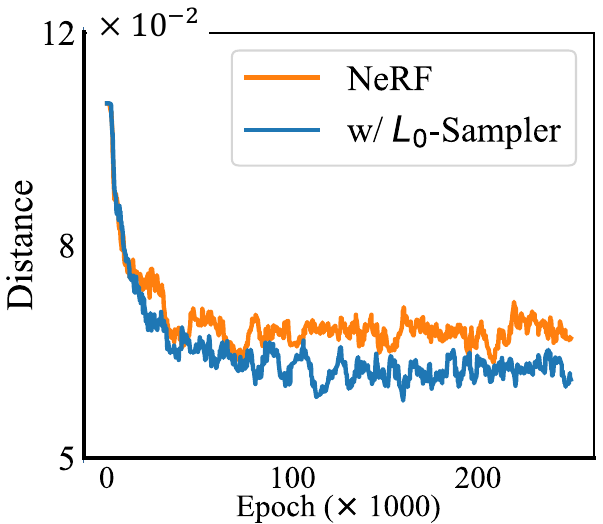}
    \caption{Sampling Distance}
    \label{fig:distance}
  \end{subfigure}
  \caption{$(a)$ Left: Coarse rendering in HVS. Middle: Fine rendering in HVS. Right: Rendering using a well-trained density network to guide sampling. $(b)$ Rendering loss comparison of NeRF and NeRF with ground truth mesh for accurate sampling. $(c)$ Average distance between sampling points and ground truth mesh during training, showing our accelerated convergence of sampling towards real geometry.}
\label{fig:intro}
\end{figure}

The advent of Neural Radiance Fields (NeRF)~\cite{nerf} has revolutionized the field of inverse rendering, providing powerful solutions for tasks like novel view synthesis, 3D surface reconstruction~\cite{neus,volumesdf}, and dynamic deformation~\cite{nerfies,dnerf}. Volume rendering plays a crucial role in the success of NeRF, as it optimizes the density and color networks to calculate pixel colors. This involves tracing rays through pixels and sampling points along the rays. By leveraging the volume rendering formula, NeRF combines the features of these sampled points to determine the final color. We are aware that in most cases, most sampled points are unoccupied and have little influence on the final result. As a result, the colors obtained through volume rendering mainly rely on points near the surface. As illustrated in \cref{fig:guide,fig:losses}, providing an accurate geometry to guide the sampling process can significantly improve the training speed and final convergence results. Therefore, a key research problem to further improve volume rendering methods such as NeRF is to further improve the sampling efficiency and concentrate the sampling points as close to the surface as possible.

NeRF introduces the Hierarchical Volume Sampling (HVS) methodology, inspired by~\cite{raytracing}, as an efficient approach for sampling. HVS utilizes volume densities to generate a Probability Density Function (PDF) at the coarse stage of each ray. This PDF guides the fine sampling process, leading to improved rendering quality, as illustrated in \Cref{fig:teaser}. In the HVS of NeRF, PDFs are approximated using piecewise constant functions, resulting in a relatively coarse estimation. No matter how precise the weights of the coarse stage can be, the sampling remains uniformly distributed within a specific interval. Although there has been a lot of works to improve its sampling process since NeRF was proposed, currently HVS is still the most stable and versatile, and therefore the most widely used sampling strategy. Considering the wide use of HVS, its further improvements will benefit volume rendering-based neural rendering and related tasks like 3D reconstruction.

In this paper, we propose \textit{$L_0$-Sampler}, which further improves the sampling process of the HVS method. Our key insight is quite straightforward: when a ray intersects a surface, the volume rendering weight function $w(t)$ of points along the ray are primarily 0, except for very few points around the surface. This behavior is analogous to the $L_0$ distance between space points and the surface. Therefore, by approximating the weight function to the indicator function, we can make the obtained sampling points approximate the potential object surface as quickly as possible, thus accelerating the training speed and final training results. Existing methods~\cite{plenoctrees, glass} proposed to apply sparsity loss to the estimated opacity values, such that the values of points around the surface are large and the values of points are close to zero. Our proposed method is completely different from these sparsity loss term based methods as we directly utilize the $L_0$ model to guide the selection of sampling points rather than applying a sparsity loss to the density distribution. Our results shown in \cref{tab:sparsity} also confirm the superiority of our proposed method.

To achieve this target, we propose to construct suitable base functions to interpolate the weight function. Through a comprehensive study, we find that piecewise exponential functions shown in \cref{fig:teaser} serve as highly suitable base functions. Specifically, they possess the ability to adaptively adjust the gradient within their intervals based on the weight being interpolated, resulting in stable and effective performance across various tasks. As illustrated in \cref{fig:distance}, the utilization of the $L_0$-Sampler approach brings the sampling points closer to the actual surface.

By applying $L_0$-Sampler to different NeRF frameworks including NeRF~\cite{nerf}, NeRF++~\cite{nerfpp}, Instant NGP~\cite{instantngp}, mip-NeRF\cite{mipnerf} and NeRF based surface reconstruction NeuS~\cite{neus}, we have observed stable performance improvements, demonstrating its adaptability across diverse datasets and techniques. In addition, one of its particularly important characteristics in practical applications is that its implementation is quite simple and parameter-free. It only requires around ten lines of code to transition from the HVS of NeRF to our method, and each step has a closed-form solution without introducing extra computational overhead. In summary, our main contributions include the following aspects:
\begin{itemize}
\item We propose the $L_0$-Sampler, an enhanced sampling strategy that concentrates sampling by shaping $w(t)$ to approximate the $L_0$ distance form.
\item We analyze the required properties to the interpolation base functions and utilize the piecewise exponential function to interpolate a quasi-$L_0$ $w(t)$.
\item Our $L_0$-Sampler can stably improve the performance of image rendering and surface reconstruction, and it is parameter-free and can be easily implemented without extra computational overhead.
\end{itemize}
\section{Related Work}
\label{relatedwork}
\noindent{\bf{Volume Rendering \& Surface Rendering.}} 
Differentiable rendering mainly includes two types: surface rendering and volume rendering. Surface rendering, exemplified by approaches like DVR~\cite{DVR} and IDR~\cite{idr}, optimizes surfaces based on multi-view images, focusing on radiance determination and often employing implicit gradients. Among these, the signed distance function (SDF) is a popular surface representation, extensively utilized in numerous studies~\cite{deepsdf, igr, idr}. Volume rendering techniques~\cite{nv, nerf} incorporate both density and color fields, effectively rendering semi-transparent materials but lacking in precise surface definition. This method computes pixel color via a weighted sum along a ray~\cite{raytracing}, with sampling crucial in weight determination. Prior studies~\cite{volume_filter} have addressed the complexities of sampling and interpolating density functions in 3D spaces. Recent efforts, including~\cite{unisurf, nddf}, have aimed to merge volume and surface rendering to balance rendering quality and geometric accuracy. Our work aligns with these efforts by modifying the sampling weight function $w(t)$ to better highlight surface features.

\noindent{\bf{Neural Radiance Field.}} Neural Radiance Field (NeRF)~\cite{nerf} has significantly impacted view synthesis and depth estimation, inspiring a wealth of enhancements~\cite{pixelnerf, tensorf} to its rendering quality and speed. Various hybrid representations like voxel grids in DVGO~\cite{dvgo} and point clouds in Point-NeRF~\cite{pointnerf} have been investigated for efficiency gains, alongside specialized structures such as hash grids in Instant NGP~\cite{instantngp}, relu fields in ReLU Fields~\cite{relufield}, sparse grids in Plenoxels~\cite{Plenoxels}, and proposal networks in mip-NeRF 360~\cite{mip360} to accelerate training.

NeRF's versatility extends to multiple domains: representing human figures~\cite{headnerf, nerfblendshape, intrinsicngp}, surface reconstruction leveraging SDF similarities~\cite{unisurf, volumesdf, neus}, dynamic scene modeling~\cite{nerfies, nonrigid, du2021nerflow, guo2021adnerf, dnerf}, deformation tasks~\cite{narf, nerfies, hypernerf}, and even relighting~\cite{NeRFactor, nerfrv}. It adapts to unbounded scenes~\cite{nerfpp, mip360} with neural networks for background modeling. The proliferation of NeRF methodologies has led to the creation of comprehensive code frameworks like Nerfstudio~\cite{nerfstudio}, NeRF-Factory~\cite{nerffactory}, Kaolin-Wisp~\cite{kaolin}, and NerfAcc~\cite{nerfacc}, which integrate many advanced algorithms.

\noindent{\bf{NeRF Sample Strategy Improvement.}} The hierarchical volume sampling (HVS) proposed by NeRF, inspired by ~\cite{efficient_sample}, has been a significant contribution to improving the sampling effects. Subsequent works have further enhanced the sampling approach from various perspectives. One aspect focuses on efficiently sampling rays, as demonstrated in papers such as~\cite{efficient_rays, few_rays}. These methods employ depth maps or context-based probability distributions to guide the selection of ray shooting locations, optimizing the sampling process. Additionally, several works utilize auxiliary networks to help determine the positions of sample points. Examples include~\cite{nerf_in_detail, donerf, terminerf, adanerf, neusample}. These networks play a crucial role in improving the accuracy and efficiency of sample point selection. Novel presentations have also been designed to enhance the NeRF sampling. For instance, mip-NeRF~\cite{mipnerf} samples conical frustums along the rays instead of individual points to reduce aliasing effects. DDNeRF~\cite{ddnerf} fits a Gaussian distribution within each interval to learn a more precise representation of the density distribution along the rays. Furthermore, certain works leverage the properties of each ray to avoid excessive sampling. Notably, the light field series of works~\cite{lightfield, lightfieldnet, lumigraph} only sample once per ray and do not rely on density during rendering. Similarly, ~\cite{autoint} proposes an automatic integration framework for calculating ray integrals, while ~\cite{reparameterized} constructs a reparameterized Monte Carlo color approximation to select points.
\section{Background and Motivation}
\label{motivation}

\begin{figure*}[htbp]
\centering
\includegraphics[width=1.0\linewidth]{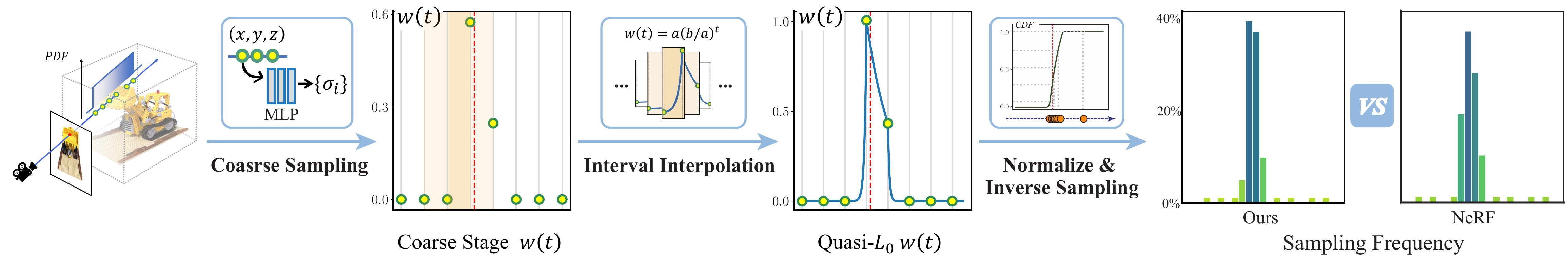}
\caption{\textbf{An overview of our $L_0$-Sampler pipeline.} The red dash line represents the surface. During hierarchical volume sampling, we first uniformly sample some points as NeRF in the coarse stage, and then through piecewise interpolation (\cref{eq:exp}), we fit a quasi-$L_0$ $w(t)$ resembling an indicator function, which is in line with the $L_0$ distance between points and surface. After normalization (\cref{eq:normalize}), it can be used as a PDF to guide inverse transform sampling (\cref{eq:sample}). The sampling frequency in each interval (right) shows that our method can make the sampling more focused near the surface.} 
\label{fig:method}
\end{figure*}
Given a point $\mathbf{p}$ and a set $S$, $\mathbf{p}, S \in \mathbb{R}^n$, the $L_0$ distance between $\mathbf{p}$ and $S$ is defined as follows:
\[d(\mathbf{p}, S) = 
\begin{cases}
  1 & \text{if } \mathbf{p} \notin S \\
  0 & \text{if } \mathbf{p} \in S.
\end{cases}\]
A distinctive characteristic of this metric is its discontinuous nature, exhibiting an abrupt transition when $\mathbf{p}$ crosses the surface $S$.
Shifting our focus to Neural Radiance Fields (NeRF)~\cite{nerf}, NeRF learns to map each point $\mathbf{p}$ and direction vector $\mathbf{d}$ in 3D space to a view-dependent radiance $c(\mathbf{p}, \mathbf{d})$ and a view-independent density $\sigma(\mathbf{p})$. The expected color $C(\mathbf{r})$ of a camera ray $\mathbf{r}(t) = \mathbf{o} + t\mathbf{d}$, constrained by the bounds $t_n$ and $t_f$, is computed as:
\begin{equation}
    \begin{split}
        C(\mathbf{r})&=\int_{t_{n}}^{t_{f}}  w(t) \mathbf{c}(\mathbf{r}(t), \mathbf{d}) dt, \\
        \text {where  } w(t)&=\sigma(\mathbf{r}(t))\exp \left(-\int_{t_{n}}^{t} \sigma(\mathbf{r}(s)) d s\right).%\nonumber
    \end{split}
\end{equation}
Here, $w(t)$ represents the contribution of the point color at $\mathbf{r}(t)$ to the cumulative color of the ray. If a distinct surface $S$ is present, the density $\sigma(\mathbf{p})$ is generally negligible for most points in space, only significantly increasing when approaching the surface. Consequently, $w(t)$ presents the following form:
\[w(t)=
\begin{cases}
  1 & \text{if } \mathbf{r}(t) \in S \\
  0 & \text{otherwise}.
\end{cases}\]
The weight function $w(t)$ displays a binary-like behavior that closely resembles the $L_0$ distance between the point $\mathbf{r}(t)$ and the surface $S$. We define this correspondence as the $\boldsymbol{L_0}$ \textbf{property} of $w(t)$. This property presents the interactions of light with surfaces, aligning well with the physics of real-world rendering.

The weight function $w(t)$ is crucial not only for rendering but also for guiding efficient sampling. Observing that points with $w = 0$ have a minor impact on the final rendering, computing values at these points is a redundant operation. It is thus more efficient to target sampling in regions close to the surface, where $w$ is much larger. This can be achieved by normalizing $w(t)$ into a probability density function (PDF) and using it for inverse transform sampling. However, obtaining a continuous representation of $w(t)$ is impractical. Instead, we turn to a coarse-to-fine estimation strategy. Initially, points $\{t_i\}$ are sampled uniformly along the ray, and their weights $\{w_i\}$ are predicted using the outputs of the coarse network as follows:
\begin{equation}
\alpha_i=1-\exp\left(-\sigma_i\delta_i\right),\qquad w_{i} =\alpha_i\prod_{j=0}^{i-1}\left(1-\alpha_j\right).
\end{equation}
Where $\delta_i = t_{i+1} - t_i$ represents the interval length. Subsequently, we extend the weights $\{w_i\}$ to get a function $w(t)$ continuously defined along the entire ray. By ensuring $w(t)$ possesses the desired $L_0$ property, we focus the majority of our sampling points at key locations, thereby enhancing efficiency in the fine sampling phase. A naive solution is to take: 
\[w(t)=
\begin{cases}
  1 & \text{if } t \text{ is near } t^* := \arg{\max{\{w_i\}}}\\
  0 & \text{otherwise}.
\end{cases}\]
While this $w(t)$ guarantees the $L_0$ property and focuses the sampling positions, it may lead to significant errors if the focus deviates from the true surface, which often happens in the early stages of training. And as discussed in \cref{sec:intro}, the HVS of NeRF extends $\{w_i\}$ into a piecewise constant $w(t)$, uniformly sampling within each interval, resulting in a weak $L_0$ property. To address this challenge, we adopt piecewise interpolation to obtain a \textbf{quasi-$L_0$ $w(t)$}. This technique balances the optimization of the residual space while preserving the $L_0$ property to a certain degree. The specifics of this interpolation technique and its role in achieving a quasi-$L_0$ $w(t)$ will be elaborated in the following section.
\section{Method}
\label{method}
We now turn to a single ray $\mathbf{o} + t\mathbf{d}$. As discussed above, after obtaining $\{w_i\}$ in the coarse stage, we seek a $w(t)$ with quasi-$L_0$ property for fine-stage sampling that satisfies $w(t_i)=w_i$ and enhances sampling efficiency. Unlike NeRF, which uses interval midpoints for $\{w_i\}$, our method interpolates $w(t)$ using the weights at interval endpoints. The integral of $w(t)$ over $\left[t_i, t_{i+1}\right]$ satisfies:
\begin{equation}
\int_{t_i}^{t_{i+1}} w(t)dt = (t_{i+1}-t_i)\int_0^1 w((t_{i+1}-t_i)s + t_i)ds.
\end{equation}
Since $\{t_i\}$ is uniformly sampled, the factor $\left(t_{i+1}-t_i\right)$ can be disregarded after normalization. Our analysis, therefore, centers on functions within $\left[0,1\right]$. After being transformed back and combined, they collectively establish the comprehensive $w(t)$ along the ray. In \cref{properties}, we explore the key properties necessary for effective interpolation. Based on that, in \cref{oursolution}, we will give our solution.

\subsection{Interpolation Techniques}
\label{properties}
Our attention turns to the interval $\left[t_i, t_{i+1}\right]$ and its mapping to the normalized interval $\left[0, 1\right]$. Here, we introduce the transformed weight function $\hat{w}(s)$, defined as $\hat{w}(s) := w\left(\left(t_{i+1} - t_i\right)s + t_i\right)$, with the boundary values $\hat{w}(0)=a$ and $\hat{w}(1)=b$, constrained by $0 \leq a, b \leq 1$ as a weight. The purpose is to interpolate $\hat{w}(s)$ within this unit interval in a manner that imitates a quasi-$L_0$ property to enhance sampling efficiency. We will enumerate and discuss the key properties that these interpolating functions should possess to meet our optimization requirement.

\textbf{Property \uppercase\expandafter{\romannumeral 1}: Capable of Precise Interpolation.} Formally, $\hat{w}(s)$ should satisfy $\hat{w}(0)=a$ and $\hat{w}(1)=b$.

\textbf{Property \uppercase\expandafter{\romannumeral 2}: Monotonic Within the Interval.} This property biases sampling towards the interval end with the greater weight. Considering the unknown location of the maximum density point within the interval, we prefer sampling to converge on the endpoints for consistency. Besides, the monotonic nature of $\hat{w}(s)$ ensures that the integrated weight across any interval $\left[t_i, t_{i+1}\right]$ is at least as great as the minimum weight at the endpoints, \ie:
\begin{equation}
\int_{t_i}^{t_{i+1}} w(t) dt = \int_0^1 \hat{w}(s) ds \geqslant \min\{a, b\}.
\end{equation}
Therefore, intervals with higher endpoint weights are more likely to be sampled after normalized, making sampling more concentrated towards them.

\begin{figure}[ht]
\begin{centering}
  \begin{subfigure}{0.45\linewidth}
    \includegraphics[width=1.\linewidth]{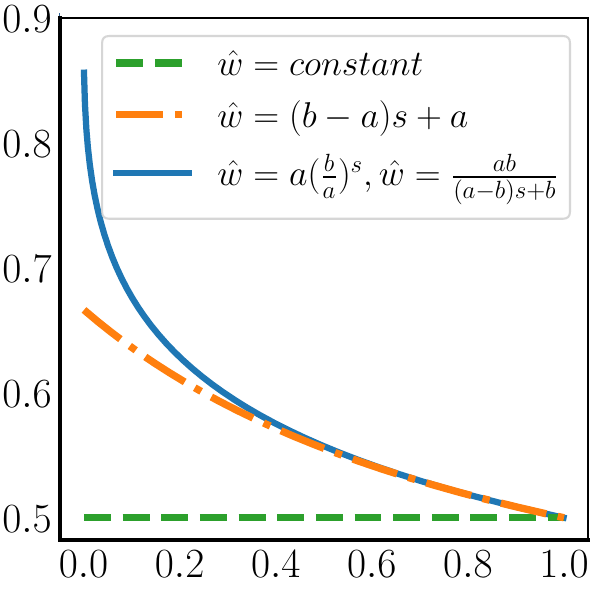}
    \caption{$bias(\hat{w})$}
    \label{fig:bias}
  \end{subfigure}
  \hfill
  \begin{subfigure}{0.45\linewidth}
    \includegraphics[width=1.\linewidth]{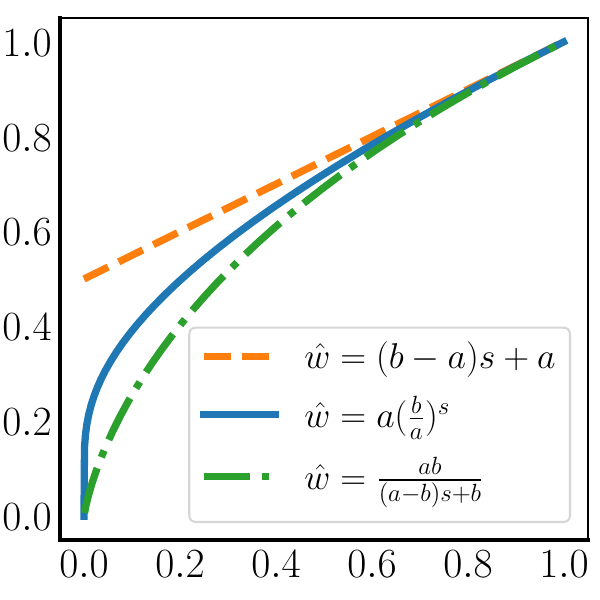}
    \caption{$\int_0^1\hat{w}(s)ds$}
    \label{fig:int}
  \end{subfigure}
\end{centering}
\caption{$(a)$ Bias of some base functions $\hat{w}$. $(b)$ Their integrals over the interval $\left[0,1\right]$. Refer to \cref{fig:functions} for function shapes.}
\label{fig:pro3}
\end{figure}
\textbf{Property \uppercase\expandafter{\romannumeral 3}: Biased Towards the Larger-weight Side}. This property is important in providing our function with quasi-$L_0$ property, especially as training progresses. Within any small segment $ds$, the sampling probability at point $s$ is $\left(\hat{w}(s)ds\right)/\left(\int_0^1\hat{w}(s)ds\right)$. Assume that $a \leqslant b$, sampling should be skewed towards $s=1$ where $\hat{w}(s)$ is higher. To measure this bias, we introduce a weighting function $f(s)$ that increases from 0 to 1 across the interval. A linear weight function, $f(s)=s$, is a straightforward choice, generating the bias metric:
\begin{equation}
   bias(\hat{w}):=\frac{\int_0^1f(s)\hat{w}(s)ds}{\int_0^1\hat{w}(s)ds} = \frac{\int_0^1s\cdot \hat{w}(s)ds}{\int_0^1\hat{w}(s)ds}.
\end{equation}
This metric, in fact, is the barycenter of the area under the curve $\hat{w}(s)$ over the interval $\left[0, 1\right]$. A larger $bias(\hat{w})$ means the $\hat{w}(t)$ has a stronger $L_0$ property. In that case, functions are typically ``steep'' in shape near $s=1$. Notably, the $bias(\hat{w})$ is dependent on endpoint values $a$ and $b$. For a fixed $\hat{w}(s)$ shape with $b=1$, $bias(\hat{w})$ becomes a function of $a$. This dependency is visualized in \cref{fig:bias}.
When $a=b=1$, monotonicity makes $\hat{w}(s)=1$, centralizing the barycenter at $t = 0.5$. The contrast is most evident when $a$ is small. A large $a, b$ gap implies $b$ is likely near a surface, prompting us to shift the bias towards $s=1$. 

However, an excessive focus on the barycenter can lead to issues since it only represents the spread of sample points within a specific interval. Considering that during normalization, the sampling probability in $\left[t_i, t_{i+1}\right]$ is proportional to the integral of $w(t)$ over that interval, an excessively steep function, $\hat{w}(s)$, may results in:
\begin{equation}
\hat{w}(s) \rightarrow \min\{w_i, w_{i+1}\}.
\end{equation}
This can result in an almost uniform distribution. Besides, when the difference between $a$ and $b$ is too large, the integral within certain intervals might approximate $\min{\{a, b\}}$ as depicted in \cref{fig:int}, reducing the expected concentration of sample points in these regions. Hence, it is crucial to find a balance in the steepness of $\hat{w}(s)$ to ensure the overall sampling distribution is focused as intended.

\textbf{Property \uppercase\expandafter{\romannumeral 4}: Computationally Efficient. }The HVS strategy originally used by NeRF is computationally efficient due to the simplicity of piecewise constant functions, which are straightforward to integrate into a PDF and to use for inverse transform sampling. To preserve this efficiency, the cumulative distribution function $\hat{W}(x):=\int_{t_n}^x \hat{w}(s)ds$ must have an explicit form, and the equation $\hat{W}(x)=r$ should be easily solvable for $r \geqslant 0$. This requires that $\hat{w}(s)$ remains appropriately simple. Our experiments further indicate that simple functions are sufficient to fulfill this task.

~\\
To summarize, when selecting a quasi-$L_0$ $\hat{w}(s)$, it is important to consider the properties we discussed above. Our pipeline is illustrated in \cref{fig:method}. After determining the interpolation of $\hat{w}(s)$ within each interval, we concatenate these functions to form the PDF. Then, similar to NeRF, by normalizing and applying inverse transform sampling, we achieve more precise sampling results. In fact, the HVS of NeRF is essentially a specific case within our proposed process. It interpolates $w(t)$ using a piecewise constant function and guides sampling with it. Thus, our approach generalizes the original HVS, offering a more universally applicable method.
%%%%%%%%%%%%%%%%%%%%%%%%%%
\subsection{Proposed Solution}
\label{oursolution}

Following our analysis, the most suitable function we have found is defined as:
\begin{equation}
\hat{w}(s) = a\left(\frac{b}{a}\right)^s.
\label{eq:exp}
\end{equation}
This function is selected due to its monotonic behavior, steep gradient, and simplicity. Notably, its integral over the interval $\left[0,1\right]$ is important for normalizing $w(t)$ to get the PDF, and is calculated as follows:
\begin{equation}
s\left(\hat{w}\right) = \int_0^1 a\left(\frac{b}{a}\right)^s ds = \frac{b - a}{\ln b - \ln a}.
\label{eq:normalize}
\end{equation}
The following equation allows us to use inverse transform sampling to find the sampling position $x$ for any residual probability $r$ in the interval:
\begin{equation}
r = \int_0^x a\left(\frac{b}{a}\right)^s ds
\Rightarrow x = \frac{\ln\left[\frac{r\left(\ln b - \ln a\right)}{a} + 1\right]}{\ln b - \ln a}.
\label{eq:sample}
\end{equation}
\begin{figure}[ht]
\begin{centering}
  \begin{subfigure}{0.45\linewidth}
    \includegraphics[width=1.\linewidth]{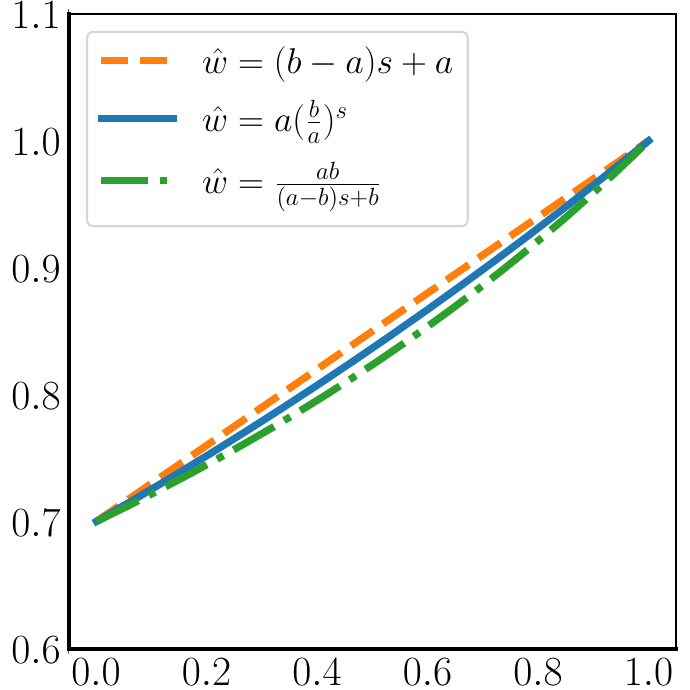}
    \caption{$a = 0.7, b = 1$}
    \label{fig:a07}
  \end{subfigure}
  % \hfill
  \begin{subfigure}{0.45\linewidth}
    \includegraphics[width=1.\linewidth]{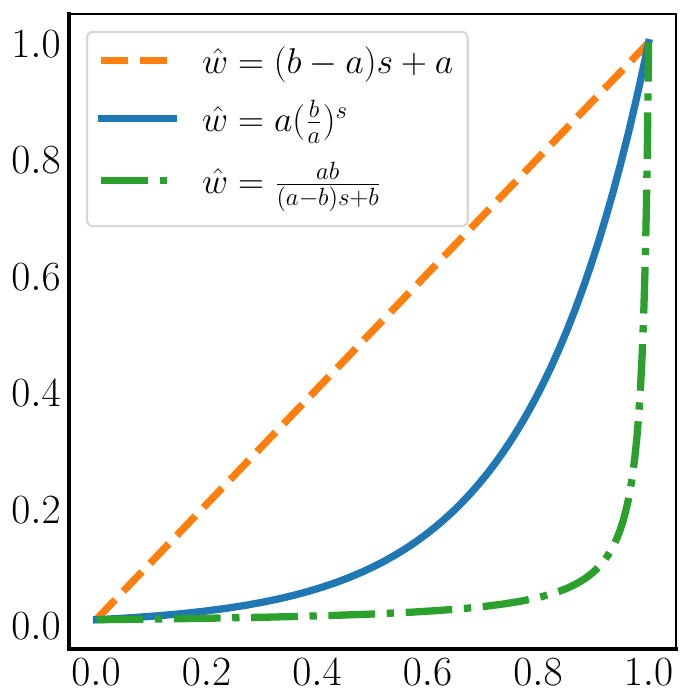}
    \caption{$a=0.01, b=1$}
    \label{fig:a001}
  \end{subfigure}
\end{centering}
\caption{\textbf{Our Solution.} They can adaptively change their shape. $(a)$ When $a$ and $b$ are similar, they approximate a linear shape. $(b)$ When $a$ is much smaller than $b$, they become steep and lead the sampling points towards the interval ends.}
\label{fig:functions}
\end{figure}
Moreover, when the values of $a$ and $b$ are relatively close, as shown in \cref{fig:a07}, this suggests the surface here is ambiguous. In these situations, our function behaves more like a linear one, promoting a more uniform sampling within that interval. Conversely, when there is a significant disparity between $a$ and $b$, exemplified in \cref{fig:a001}, the curve of the function becomes steeper, resembling an exponential form. This steepness causes its barycenter to shift towards 1, granting it a quasi-$L_0$ property, as depicted in \cref{fig:bias}. This shift results in more focused sampling in areas with clearer surfaces. Essentially, this represents an adaptive sampling strategy controlled by the ratio $b/a$.

Additionally, we also consider the piecewise inverse function: 
\begin{equation}
\hat{w}(s) = \frac{ab}{\left(a-b\right)s + b}.
\end{equation}

The function is named because it is derived from $1/s$. The equations it needs in normalization and inverse transform sampling are:
\begin{equation}
s\left(\hat{w}\right)=\int_0^1 \hat{w}(s) ds = \frac{ab}{b-a}\left(\ln b-\ln a\right)\\
\end{equation}
\begin{equation}
r = \int_0^x \hat{w}(s) ds
\Rightarrow x = \frac{b}{b-a}\left[\exp\left(\frac{r\left(b-a\right)}{ab}\right)-1\right].
\end{equation}
It exhibits similar properties to the exponential function and can often yield satisfactory results. As illustrated in \cref{fig:bias}, it has the same $bias(\hat{w})$ as $a\left(b/a\right)^s$, indicating its comparable effects within the interval. However, its performance is less consistent, possibly due to an excessive steepness demonstrated in \cref{fig:int}, which may affect the sampling probability adversely, as discussed in \cref{properties}.

\begin{figure}[ht]
\centering
\includegraphics[width=0.4\textwidth]{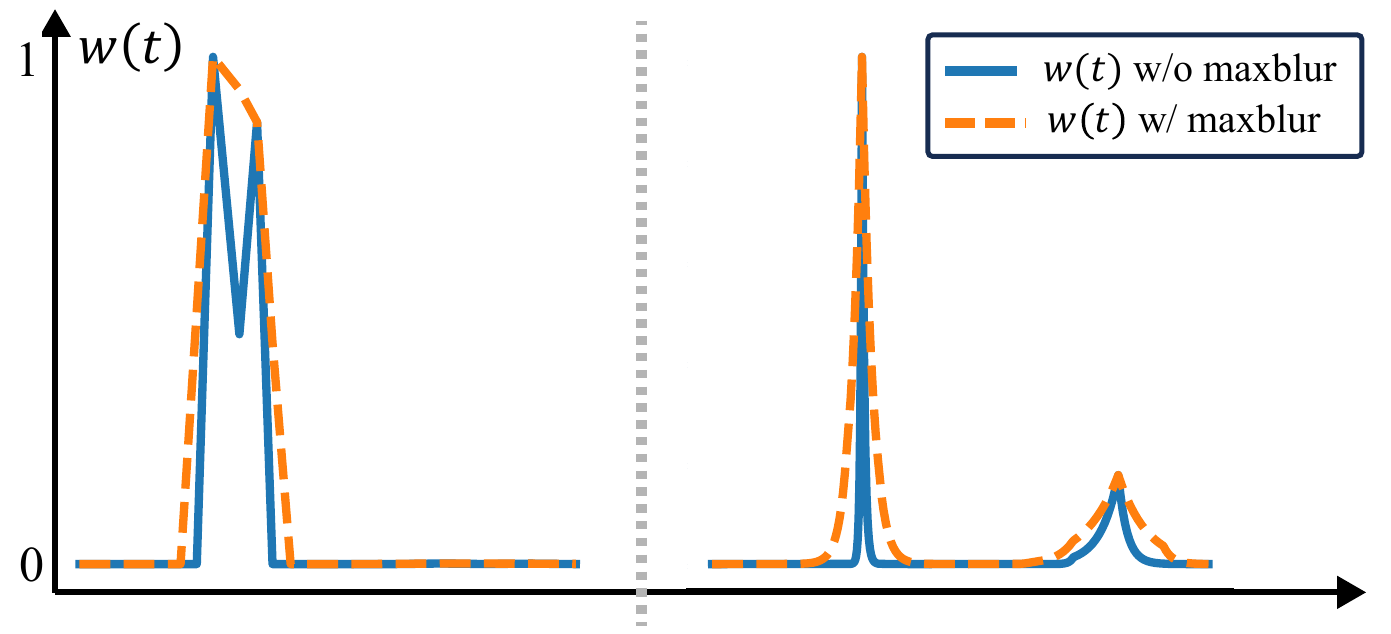}
\caption{\textbf{Effects of Maxblur.} Left: it makes $w(t)$ smoother. Right: by broadening the peak areas of $w(t)$, it enhances the probability of sampling within these intervals.}
\label{fig:maxblur}
\end{figure}

Besides, to further refine the weight distribution before sampling, we incorporate the ``maxblur'' technique from mip-NeRF~\cite{maxblur, mipnerf} into our framework:
\begin{equation}
    w_{i}^{\prime} = \frac{1}{2}\left(\max\left(w_{i-1}, w_i\right) + \max\left(w_i, w_{i+1}\right)\right)+0.01.
\end{equation}
This adjustment generates a smoother weight distribution close to reality. We find that this modification works well with our $L_0$-Sampler by broadening the peak area of $\{w_i\}$, which our steep $w(t)$ then sharpens, achieving a more balanced sampling between intervals (\cref{fig:maxblur}).
\section{Experiments}
\label{experiments}
\subsection{Experimental Settings}

\noindent{\bf{Datasets:}} We select datasets corresponding to those used in the original works. For our evaluations involving NeRF~\cite{nerf}, mip-NeRF~\cite{mipnerf}, and Instant NGP~\cite{instantngp}, we evaluate our approach using the NeRF Synthetic and LLFF dataset generated by NeRF~\cite{nerf} and LLFF~\cite{llff} respectively. In the case of NeRF++, we use scenes from the LF dataset~\cite{lfdata}, each densely populated with 2k-4k hand-held captured images, with camera parameters recovered via Structure from Motion (SfM). To evaluate the impact on NeuS, we utilize cases from the DTU dataset~\cite{dtus} and BlendedMVS dataset~\cite{bmvs}, offering a diverse range of materials, appearances, and geometries. The DTU dataset scenes each contain 49 or 64 images with a resolution of $1600 \times 1200$, while the BlendedMVS scenes are rendered at a resolution of $576 \times 768$. All scenes in the two datasets are provided with masks.

\noindent{\bf{Metrics:}} To quantitatively evaluate the rendered results on novel view synthesis, we use PSNR, SSIM~\cite{ssim} (higher is better for both), and the VGG implementation of LPIPS~\cite{lpips} (lower is better). For geometric results, we employ the Marching Cubes algorithm~\cite{mcube} to extract surfaces from the volumetric data.

\noindent{\bf{Implementation Details:}} Our implementation of $L_0$-Sampler, alongside other comparative experiments, is conducted using PyTorch on a single NVIDIA 3090 GPU. When comparing with others, we keep the training methods the same as those used in previous works, with the only difference being our new sampling technique. Notably, for the specific sampling method of mip-NeRF, which involves conical frustums, we adopt the density of each frustum as the density of its interval midpoint to enable interpolation.

\subsection{Comparisons}
\label{comparison}

\renewcommand{\arraystretch}{1.4}
\begin{table*}[tbp]
    \centering
    \bgroup
    \resizebox{\linewidth}{!}{
    \begin{tabular}{l@{\hskip 2.0pt}|@{\hskip 2.0pt}c@{\hskip 2.0pt}c@{\hskip 2.0pt}|@{\hskip 2.0pt}c@{\hskip 2.0pt}c@{\hskip 2.0pt}|@{\hskip 2.0pt}c@{\hskip 2.0pt}c@{\hskip 2.0pt}|@{\hskip 2.0pt}c@{\hskip 2.0pt}c@{\hskip 2.0pt}|@{\hskip 2.0pt}c@{\hskip 2.0pt}c@{\hskip 2.0pt}|@{\hskip 2.0pt}c@{\hskip 2.0pt}c@{\hskip 2.0pt}|@{\hskip 2.0pt}c@{\hskip 2.0pt}c@{\hskip 2.0pt}|@{\hskip 2.0pt}c@{\hskip 2.0pt}c@{}}
    \multirow{2}{*}{\textbf{Methods}} & \multicolumn{2}{c@{\hskip 2.0pt}|@{\hskip 2.0pt}}{\textbf{Lego}} & \multicolumn{2}{c@{\hskip 2.0pt}|@{\hskip 2.0pt}}{\textbf{Chair}} & \multicolumn{2}{c@{\hskip 2.0pt}|@{\hskip 2.0pt}}{\textbf{Drums}} & \multicolumn{2}{c@{\hskip 2.0pt}|@{\hskip 2.0pt}}{\textbf{Ship}}& \multicolumn{2}{c@{\hskip 2.0pt}|@{\hskip 2.0pt}}{\textbf{Ficus}}& \multicolumn{2}{c@{\hskip 2.0pt}|@{\hskip 2.0pt}}{\textbf{Materials}}& \multicolumn{2}{c@{\hskip 2.0pt}|@{\hskip 2.0pt}}{\textbf{Hotdog}}& \multicolumn{2}{c@{\hskip 2.0pt}@{\hskip 2.0pt}}{\textbf{Mic}}\\

      & PSNR$\uparrow$& LPIPS$\downarrow$ & PSNR$\uparrow$& LPIPS$\downarrow$ & PSNR$\uparrow$& LPIPS$\downarrow$ & PSNR$\uparrow$& LPIPS$\downarrow$& PSNR$\uparrow$& LPIPS$\downarrow$ & PSNR$\uparrow$& LPIPS$\downarrow$ & PSNR$\uparrow$& LPIPS$\downarrow$& PSNR$\uparrow$& LPIPS$\downarrow$\\
    \hline
    \hline
    NeRF~\cite{nerf}          & 31.39 &0.0400	&34.52	&0.0283 &25.59	&0.0741 &29.47	&0.1409 &28.92	&0.0416 &29.59	&0.0432 &36.80	&0.0298 &33.18 &0.0272\\
    w/ $L_0$-Sampler                   & \textbf{31.97} & \textbf{0.0346} & \textbf{34.92}  &  \textbf{0.0257} & \textbf{25.72} & \textbf{0.0685}  &  \textbf{29.80} & \textbf{0.1342} & \textbf{29.21} &  \textbf{0.0368} & \textbf{29.77} & \textbf{0.0386} & \textbf{37.02} & \textbf{0.0278}& \textbf{33.61} & \textbf{0.0250}\\
    \hline
    mip-NeRF~\cite{mipnerf}     &33.86	&0.0398 &33.61	&0.0407 &24.98	&0.096 &28.64&0.1899 &31.87	&0.0345 &30.21	&0.0559 &36.05 &\textbf{0.0448}	&34.00 &\textbf{0.0211}\\
    							
    w/ $L_0$-Sampler                & \textbf{34.31} & \textbf{0.0380} & \textbf{33.71} &  \textbf{0.0405} & \textbf{25.12} & \textbf{0.0939}  &  \textbf{28.67} & \textbf{0.1898} & \textbf{32.46}  &  \textbf{0.0302} & \textbf{30.34} & \textbf{0.0548} & \textbf{36.18} & 0.0452& \textbf{34.02} & 0.0214\\

    \hline
    Instant NGP~\cite{instantngp}             &32.64	&0.0900 &32.07	&0.1050 &23.68	&\textbf{0.1469} &29.04	&\textbf{0.1926} &29.51	&0.1485 &28.02	&0.2385 &34.68	&0.0658 &31.57 &0.0521	\\
    
    w/ $L_0$-Sampler             & \textbf{33.29} & \textbf{0.0726} & \textbf{33.05} &  \textbf{0.0852} & \textbf{24.05} & 0.1616  &  \textbf{29.31} & 0.1960 & \textbf{29.96}  &  \textbf{0.1363} & \textbf{28.58} & \textbf{0.2218} & \textbf{35.66} & \textbf{0.0603}& \textbf{31.96} & \textbf{0.0505}\\

    \hline
    \hline

    \end{tabular}
    }
    \egroup
\caption{{\bf Quantitative Comparison.} The table compares the performance of various NeRF-based methods to their enhanced versions using our $L_0$-Sampler on the NeRF Blender datasets. Metrics used are PSNR ($\uparrow$) / LPIPS ($\downarrow$). We change the sampling strategy in each method into our $L_0$-Sampler. In NeRF and Instant NGP, we use piecewise exponential functions, while in mip-NeRF we use piecewise inverse functions for interpolation. We observe steady improvements post \textbf{almost the same training time} across multiple datasets and tasks.}
    \vspace{-1em}
    \label{tab:quantity}
\end{table*}
\noindent{\bf{Quantitative Comparison.}}
We integrate our $L_0$-Sampler into several works using volume rendering and importance sampling, including NeRF~\cite{nerf}, mip-NeRF~\cite{mipnerf}, and the torch version of Instant NGP~\cite{instantngp, torchngp, tang2022compressible}. Results are shown in \cref{tab:quantity}. Notably, we consistently improve PSNR across datasets, and the generally lower LPIPS suggests that our method can capture more features of the input images. The outcomes demonstrate the efficacy of our method across various datasets and tasks, further indicating its broad applicability. Although the enhancements may appear modest, the novel perspective from which we update the HVS allows our method to be combined with others, as shown with mip-NeRF and Instant NGP, leading to more precise and detailed rendering results.

Furthermore, we distinguish our method from sparsity loss approaches often used in NeRF-related works. The specific expression of the loss is:
\begin{equation}
   \mathcal{L}_{\textrm{sparsity}} = \beta_{s}\frac{1}{N}\sum_{k=1}^{N}\lvert1-\exp(-\delta\sigma_k)\rvert.
\end{equation}
Here we take $\beta_{s}=0.01$ and $\delta=0.1$. While the idea behind them appears similar to our work, these approaches aim to condense the volume density, while our method concentrates on refining the sampling of points within an already determined density. The comparison results are shown in \cref{tab:sparsity}. Our method also offers the advantage of not requiring weight adjustments for each dataset or additional computations for loss evaluation during training, which improves training efficiency.

\renewcommand{\arraystretch}{1.}
\begin{table}[t]
    \begin{center}
    \resizebox{\linewidth}{!}{
    \begin{tabular}{|l|cc|cc|cc|cc|}
    \hline
                  \multirow{2}{*}{\textbf{Method} } & \multicolumn{2}{c|}{\textbf{Lego}} & \multicolumn{2}{c|}{\textbf{Chair}} & \multicolumn{2}{c|}{\textbf{Ficus}} & \multicolumn{2}{c|}{\textbf{Materials}} \\
    % \cline{2-9}
               & PSNR $\uparrow$ &LPIPS $\downarrow$ & PSNR $\uparrow$  & LPIPS $\downarrow$ &PSNR $\uparrow$ & LPIPS $\downarrow$&PSNR $\uparrow$ & LPIPS $\downarrow$  \\
    \hline
    NeRF &31.39&0.0400&34.52&0.0283&28.92&0.0416&29.59&0.0432\\
    w/ sparsity loss       &31.58&0.0378&34.53&0.0302&28.83&0.0420&29.53&0.0441\\
    w/ $L_0$-Sampler         &\textbf{31.97}&\textbf{0.0346}&\textbf{34.92}&\textbf{0.0257}&\textbf{29.21}&\textbf{0.0368}&\textbf{29.77}&\textbf{0.0386}\\
    \hline
    \end{tabular}}
    \end{center}
    \vspace{-5mm}
    \caption{{\bf Comparison with Sparsity Loss.} The loss generally improves the rendering results but is not as effective as $L_0$-Sampler.
    }
    \label{tab:sparsity}
    \vspace{-3mm}
\end{table}
\begin{figure}[t]
\footnotesize
\centering%
\mpage{0.19}{%
		\begin{overpic}[width=\linewidth]{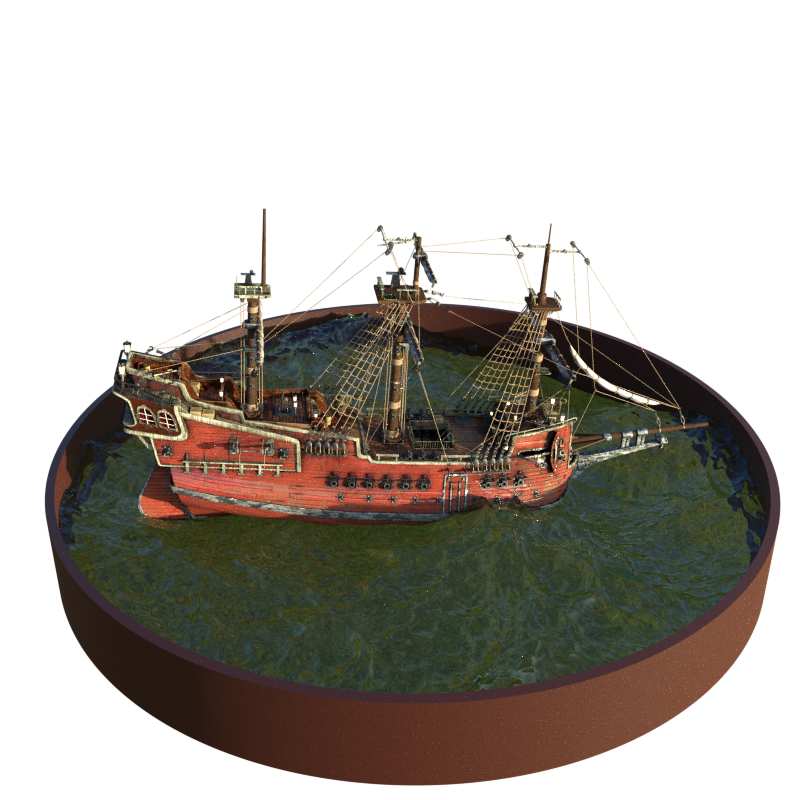}%
			\put(37, 45){\color{red}$\Box$}%
		\end{overpic}\\[-0.5mm]%
		\emph{Ship}
	}%
	\hspace{-1mm}\hfill%
	\mpage{0.8}{%
		\mpage{0.32}{%
			\includegraphics[width=\linewidth]{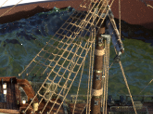}\\[0.2mm]%
		}\hspace{-1mm}\hfill%
		\mpage{0.32}{%
			\includegraphics[width=\linewidth]{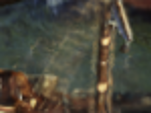}\\[0.2mm]%
		}\hspace{-1mm}\hfill%
		\mpage{0.32}{%
			\includegraphics[width=\linewidth]{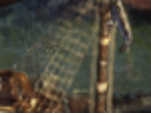}\\[0.2mm]%
		}
	}\\[1mm]%
\mpage{0.19}{%
		\begin{overpic}[width=\linewidth]{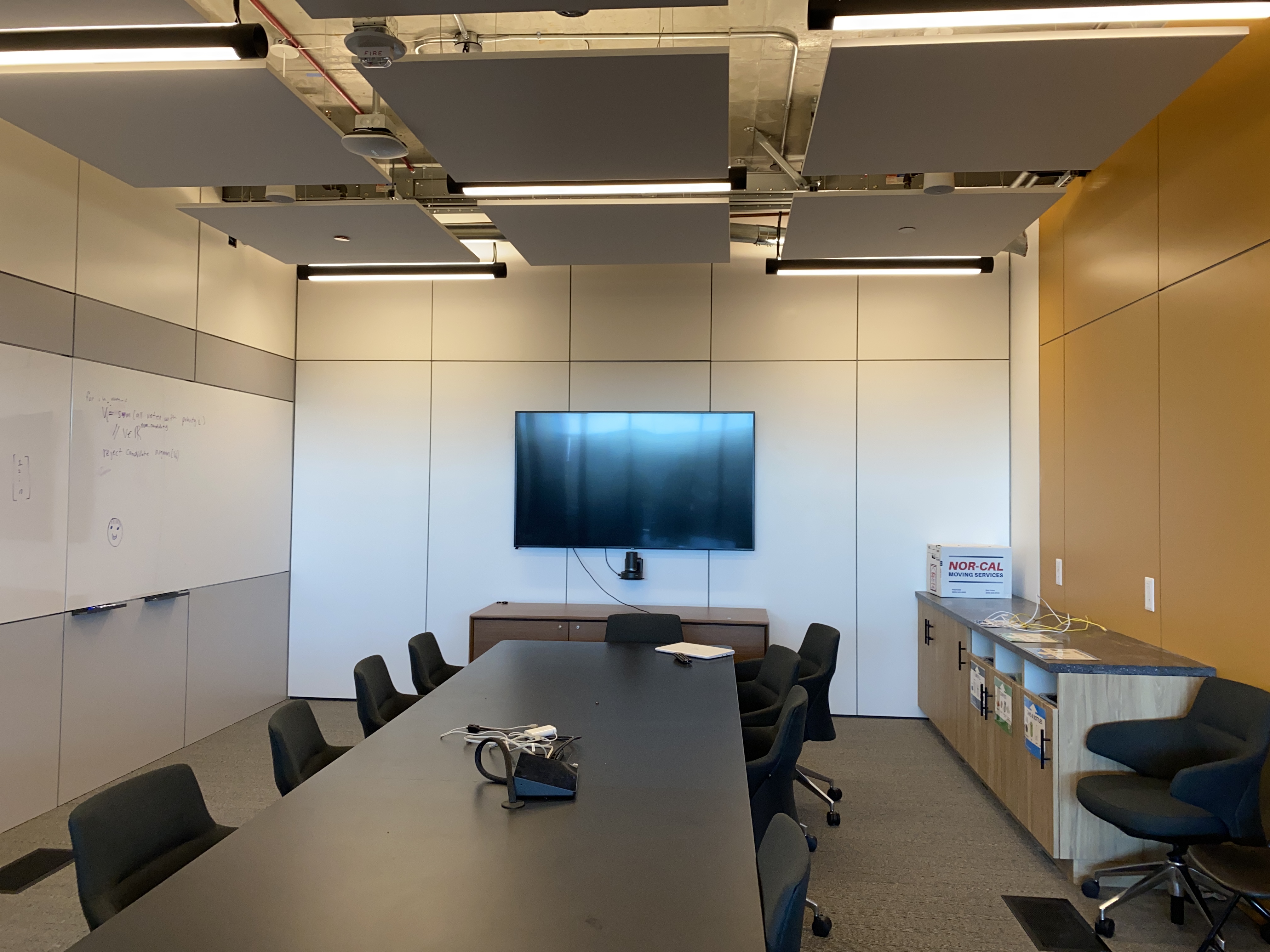}%
			\put(13, 63){\color{red}$\Box$}%
		\end{overpic}\\[-0.5mm]%
		\emph{Room}
	}%
	\hspace{-1mm}\hfill%
	\mpage{0.8}{%
		\mpage{0.32}{%
			\includegraphics[width=\linewidth]{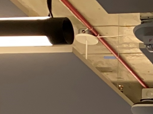}\\[0.2mm]%
		}\hspace{-1mm}\hfill%
		\mpage{0.32}{%
			\includegraphics[width=\linewidth]{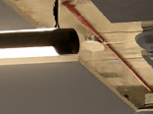}\\[0.2mm]%
		}\hspace{-1mm}\hfill%
		\mpage{0.32}{%
			\includegraphics[width=\linewidth]{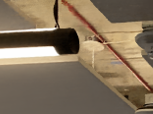}\\[0.2mm]%
		}
	}\\[1mm]%
	\mpage{0.19}{%
	}%
	\hspace{-1mm}\hfill%
	\mpage{0.8}{%
		\mpage{0.3}{%
			GT%
		}\hspace{-1mm}\hfill%
		\mpage{0.3}{%
			NeRF~\cite{nerf}%
		}\hspace{-1mm}\hfill%
		\mpage{0.3}{%
			 w/ $L_0$-Sampler%
		}
	}\\[2mm]%
%######################################################
\mpage{0.19}{%
		\begin{overpic}[width=\linewidth]{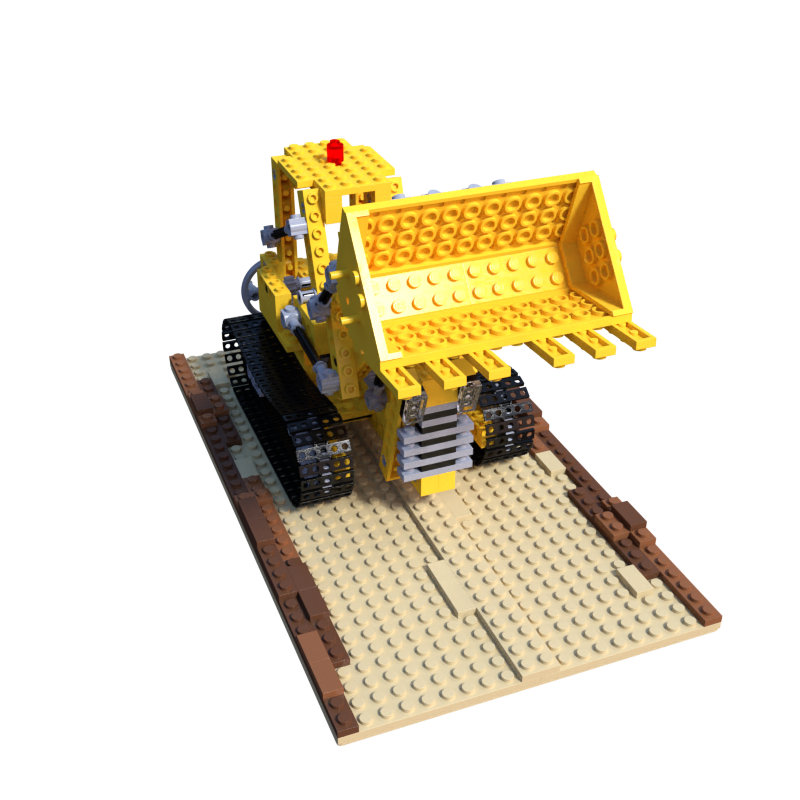}%
			\put(42, 57){\color{red}$\Box$}%
		\end{overpic}\\[-0.5mm]%
		\emph{Lego}
	}%
	\hspace{-1mm}\hfill%
	\mpage{0.8}{%
		\mpage{0.32}{%
			\includegraphics[width=\linewidth]{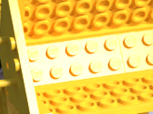}\\[0.2mm]%
		}\hspace{-1mm}\hfill%
		\mpage{0.32}{%
			\includegraphics[width=\linewidth]{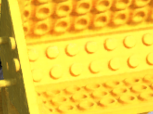}\\[0.2mm]%
		}\hspace{-1mm}\hfill%
		\mpage{0.32}{%
			\includegraphics[width=\linewidth]{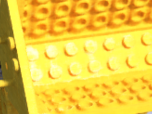}\\[0.2mm]%
		}
	}\\[1mm]%
\mpage{0.19}{%
		\begin{overpic}[width=\linewidth]{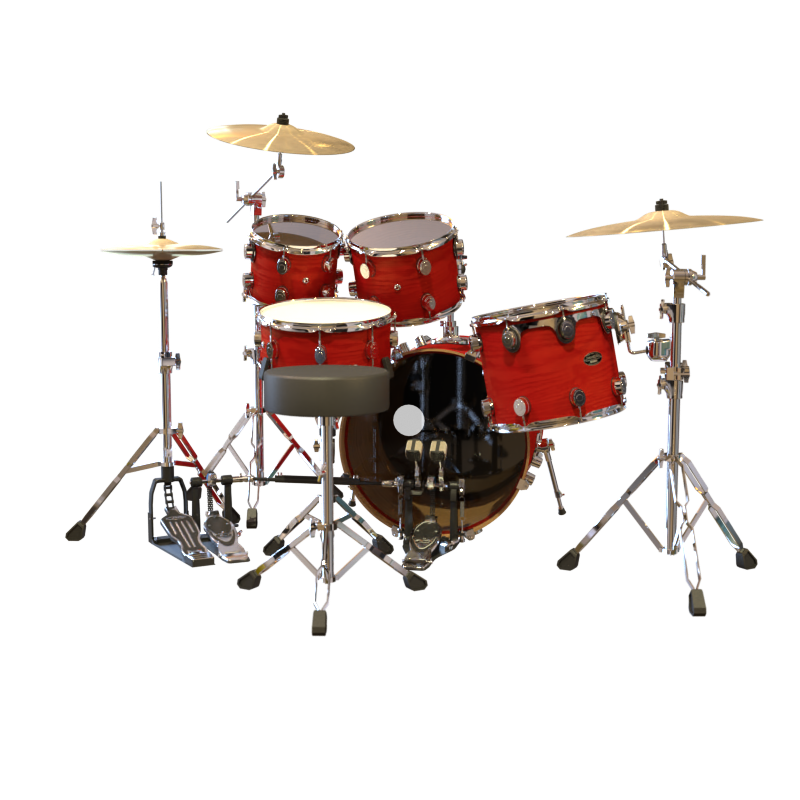}%
			\put(45, 25){\color{red}$\Box$}%
		\end{overpic}\\[-0.5mm]%
		\emph{Drums}
	}%
	\hspace{-1mm}\hfill%
	\mpage{0.8}{%
		\mpage{0.32}{%
			\includegraphics[width=\linewidth]{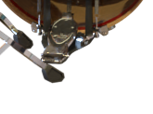}\\[0.2mm]%
		}\hspace{-1mm}\hfill%
		\mpage{0.32}{%
			\includegraphics[width=\linewidth]{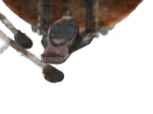}\\[0.2mm]%
		}\hspace{-1mm}\hfill%
		\mpage{0.32}{%
			\includegraphics[width=\linewidth]{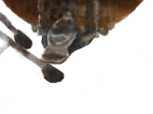}\\[0.2mm]%
		}
	}\\[1mm]%
	\mpage{0.19}{%
	}%
	\hspace{-1mm}\hfill%
	\mpage{0.8}{%
		\mpage{0.3}{%
			GT%
		}\hspace{-1mm}\hfill%
		\mpage{0.3}{%
			mip-NeRF~\cite{mipnerf}%
		}\hspace{-1mm}\hfill%
		\mpage{0.3}{%
			 w/ $L_0$-Sampler%
		}
	}\\[2mm]%
%###################################################
\mpage{0.19}{%
		\begin{overpic}[width=\linewidth]{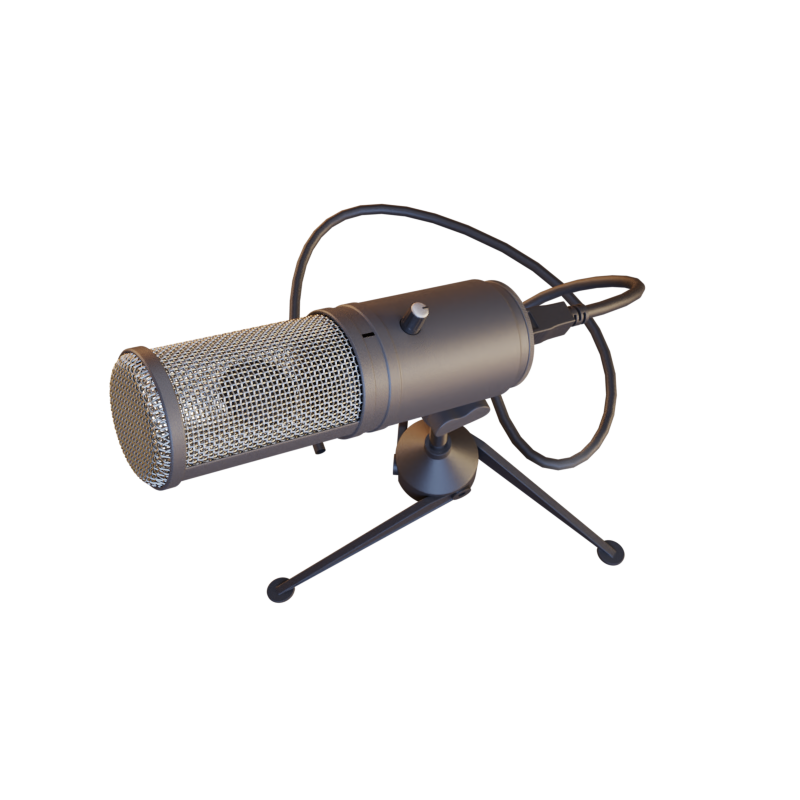}%
			\put(25, 45){\color{red}$\Box$}%
		\end{overpic}\\[-0.5mm]%
		\emph{Mic}
	}%
	\hspace{-1mm}\hfill%
	\mpage{0.8}{%
		\mpage{0.32}{%
			\includegraphics[width=\linewidth]{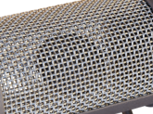}\\[0.2mm]%
		}\hspace{-1mm}\hfill%
		\mpage{0.32}{%
			\includegraphics[width=\linewidth]{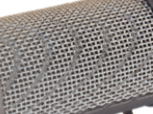}\\[0.2mm]%
		}\hspace{-1mm}\hfill%
		\mpage{0.32}{%
			\includegraphics[width=\linewidth]{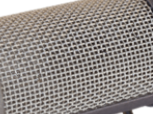}\\[0.2mm]%
		}
	}\\[1mm]%
\mpage{0.19}{%
		\begin{overpic}[width=\linewidth]{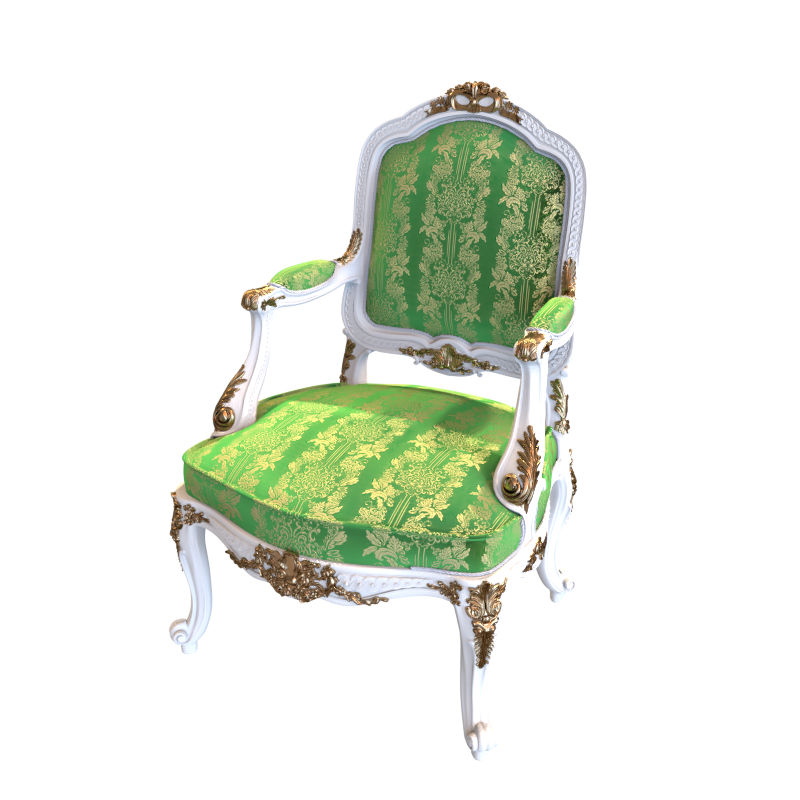}%
			\put(50, 62){\color{red}$\Box$}%
		\end{overpic}\\[-0.5mm]%
		\emph{Chair}
	}%
	\hspace{-1mm}\hfill%
	\mpage{0.8}{%
		\mpage{0.32}{%
			\includegraphics[width=\linewidth]{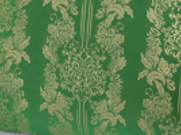}\\[0.2mm]%
		}\hspace{-1mm}\hfill%
		\mpage{0.32}{%
			\includegraphics[width=\linewidth]{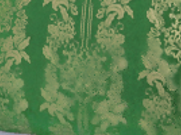}\\[0.2mm]%
		}\hspace{-1mm}\hfill%
		\mpage{0.32}{%
			\includegraphics[width=\linewidth]{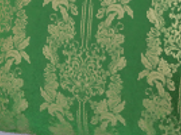}\\[0.2mm]%
		}
	}\\[1mm]%
	\mpage{0.19}{%
	}%
	\hspace{-1mm}\hfill%
	\mpage{0.8}{%
		\mpage{0.3}{%
			GT%
		}\hspace{-1mm}\hfill%
		\mpage{0.3}{%
			Instant NGP~\cite{instantngp}%
		}\hspace{-1mm}\hfill%
		\mpage{0.3}{%
			w/ $L_0$-Sampler%
		}
	}\\[2mm]%
\caption{
\tb{Qualitative Comparison.} Rendering results of different methods on NeRF-synthetic and LLFF datasets. Our method shows higher quality in rendering details.
}
\label{fig:syn}
\vspace{-5mm}
\end{figure}

\begin{figure}[ht]
\centering

\begin{subfigure}{0.15\textwidth}
    \centering
    % \vfill
    \includegraphics[width=\linewidth]{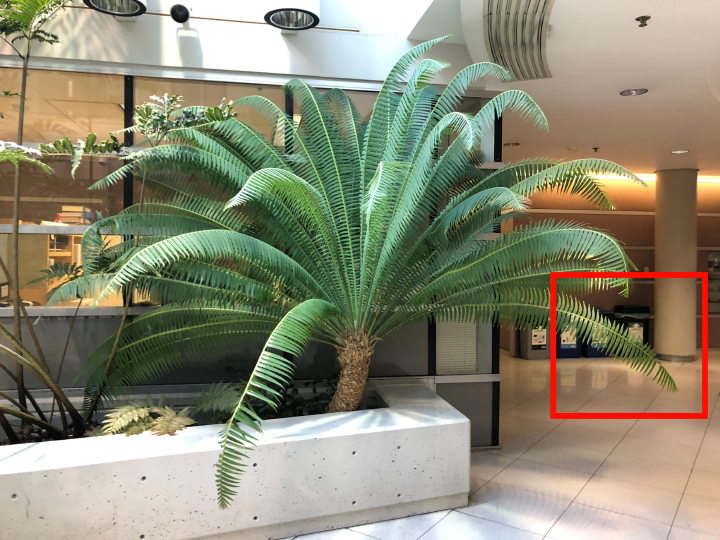}
    % \vfill
    \caption{Reference}
    \label{fig:depthrgb}
  \end{subfigure}
\begin{subfigure}{0.15\textwidth}
    \centering
    \includegraphics[width=\linewidth]{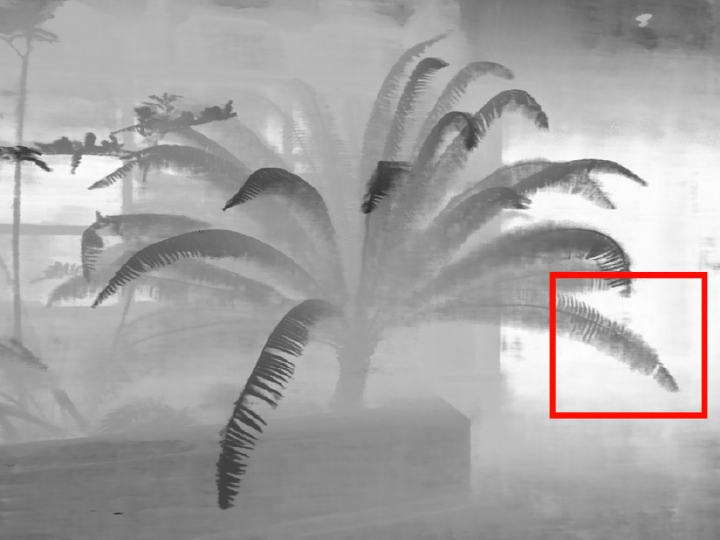}
    \caption{mip-NeRF~\cite{mipnerf}}
    \label{fig:depthsubfigure1}
  \end{subfigure}
  % \hfill
  \begin{subfigure}{0.15\textwidth}
    \centering
    \includegraphics[width=\linewidth]{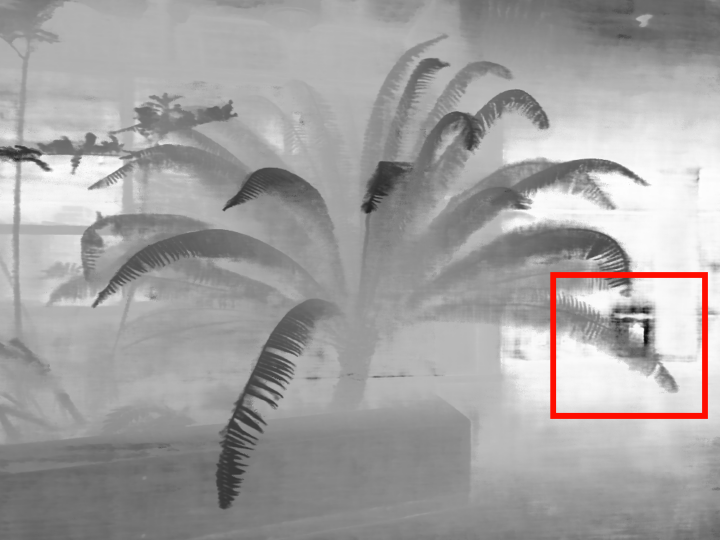}
    \caption{w/ $L_0$-Sampler}
    \label{fig:depthsubfigure2}
  \end{subfigure}

\caption{Depth maps of mip-NeRF with 32 sample points per ray. Our $L_0$-Sampler can help it to capture more geometry details.}
\label{fig:depth}
\end{figure}

\begin{figure}[ht]
\centering
\begin{subfigure}{0.206\textwidth}
    \centering
    \includegraphics[width=\linewidth]{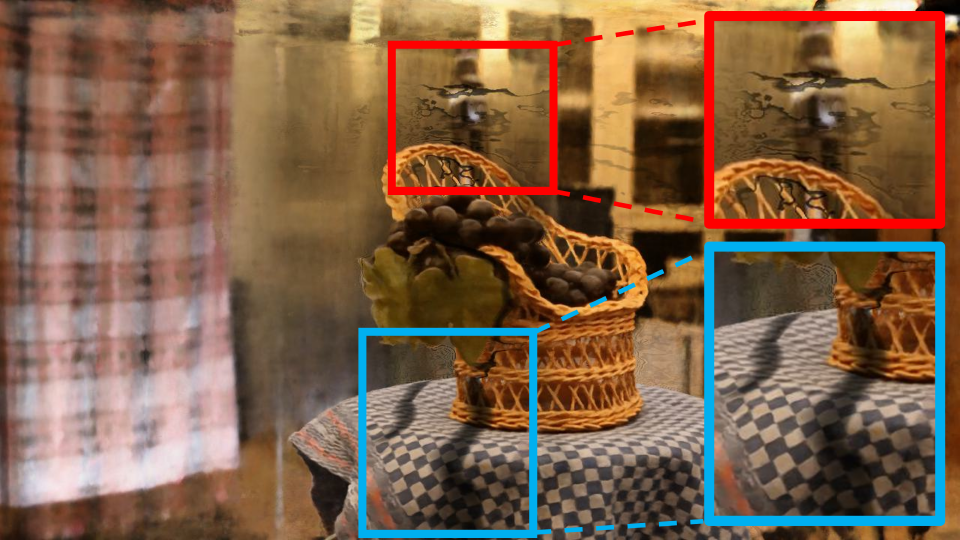}
    \caption{NeRF++}
    % \label{fig:depthsubfigure1}
  \end{subfigure}
\begin{subfigure}{0.055\textwidth}
    \includegraphics[width=\linewidth]{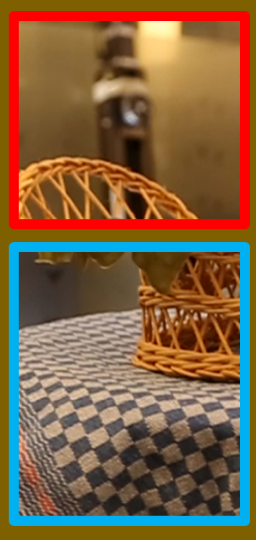}
    \caption{GT}
    % \label{fig:depthsubfigure1}
  \end{subfigure}
  % \hfill
  \begin{subfigure}{0.206\textwidth}
    \centering
    \includegraphics[width=\linewidth]{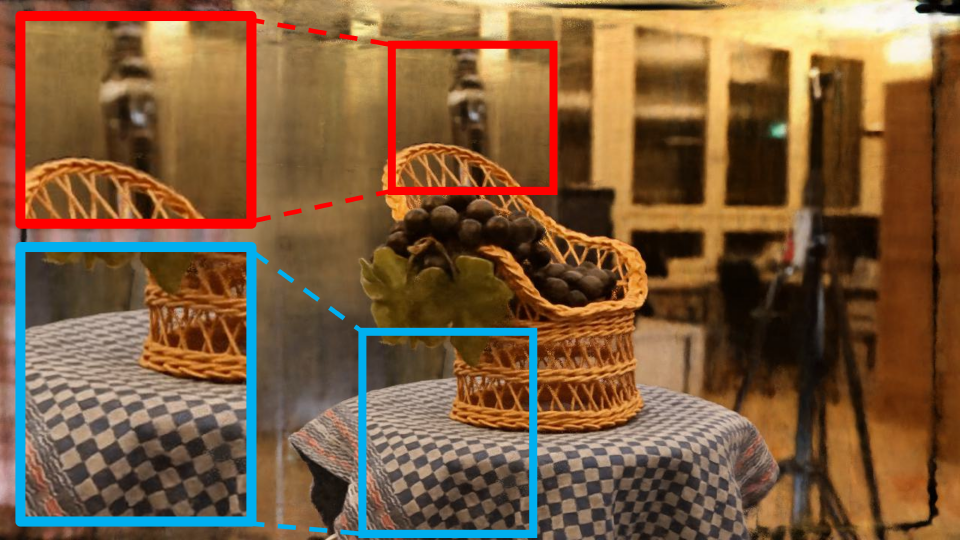}
    \caption{w/ $L_0$-Sampler}
    % \label{fig:depthsubfigure2}
  \end{subfigure}
\caption{$L_0$-Sampler can alleviate the artifacts in some views.}
\label{fig:real}
\end{figure}

\noindent{\bf{Qualitative Comparison.}}
\cref{fig:syn} provides a qualitative comparison between NeRF, mip-NeRF, and Instant NGP, and their enhanced versions utilizing our $L_0$-Sampler on the NeRF-Synthetic and LLFF datasets. It is evident that our $L_0$-Sampler helps to capture challenging details, notably highlights, complex textures, and thin structures. And like depth maps in \cref{fig:depth} shows, our sampling captures more geometric details, especially under conditions of sparse sampling points. Furthermore, \cref{fig:real} depicts the improvement brought about by $L_0$-Sampler on NeRF++~\cite{nerfpp} on Basket case in the LF dataset. Our method enhances the rendering quality of real scenes and reduces rendering artifacts. 

\begin{figure}[ht]
\centering
%\framebox[4.0in]{$\;$}
\mpage{0.19}{%
		\begin{overpic}[width=\linewidth]{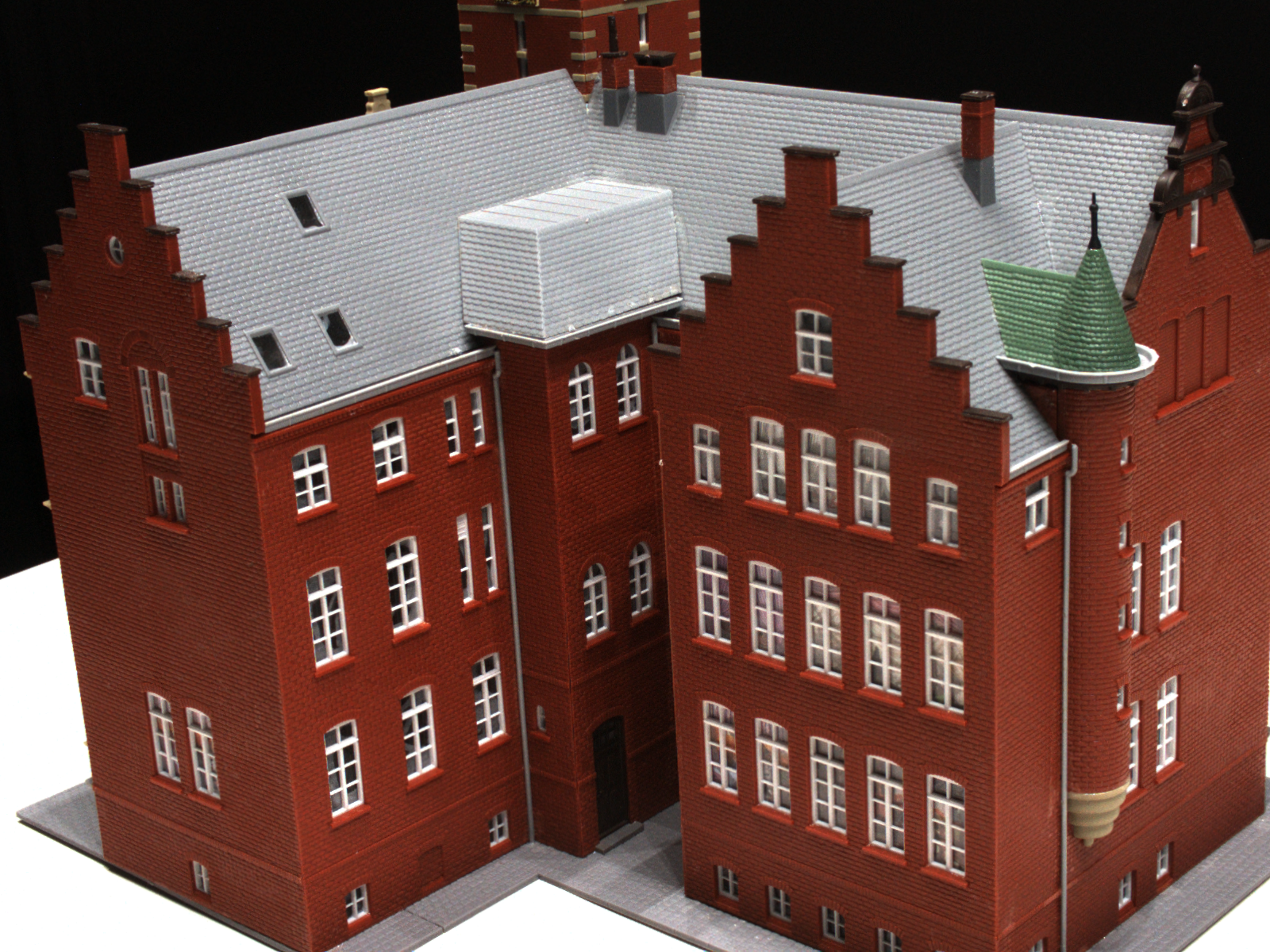}%
			\put(40, 60){\color{red}\bm{$\Box$}}
                \put(80, 0){\color{yellow}\bm{$\Box$}}%
		\end{overpic}\\[-0.5mm]%
		\emph{DTU24}
	}%
	\hspace{-1mm}\hfill%
	\mpage{0.8}{%
		\mpage{0.32}{%
			\includegraphics[width=\linewidth]{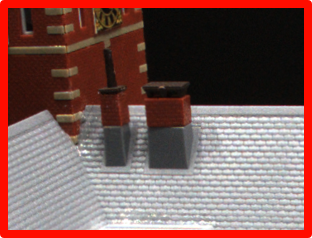}\\[0.2mm]%
		}\hspace{-1mm}\hfill%
		\mpage{0.32}{%
			\includegraphics[width=\linewidth]{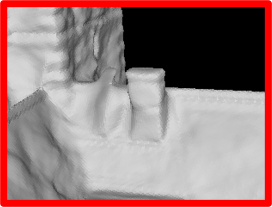}\\[0.2mm]%
		}\hspace{-1mm}\hfill%
		\mpage{0.32}{%
			\includegraphics[width=\linewidth]{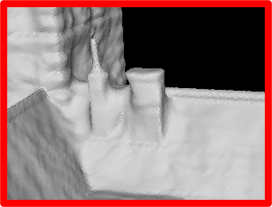}\\[0.2mm]%
		}
	}\\[1mm]%
	\mpage{0.19}{%
	}%
	\hspace{-1mm}\hfill%
        \mpage{0.8}{%
		\mpage{0.32}{%
			\includegraphics[width=\linewidth]{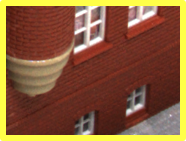}\\[0.2mm]%
		}\hspace{-1mm}\hfill%
		\mpage{0.32}{%
			\includegraphics[width=\linewidth]{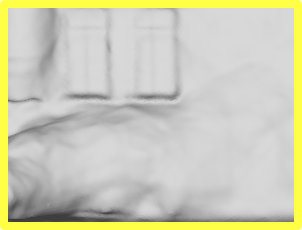}\\[0.2mm]%
		}\hspace{-1mm}\hfill%
		\mpage{0.32}{%
			\includegraphics[width=\linewidth]{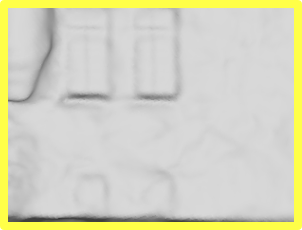}\\[0.2mm]%
		}
	}\\[1mm]%

%######################################################
\mpage{0.19}{%
		\begin{overpic}[width=\linewidth]{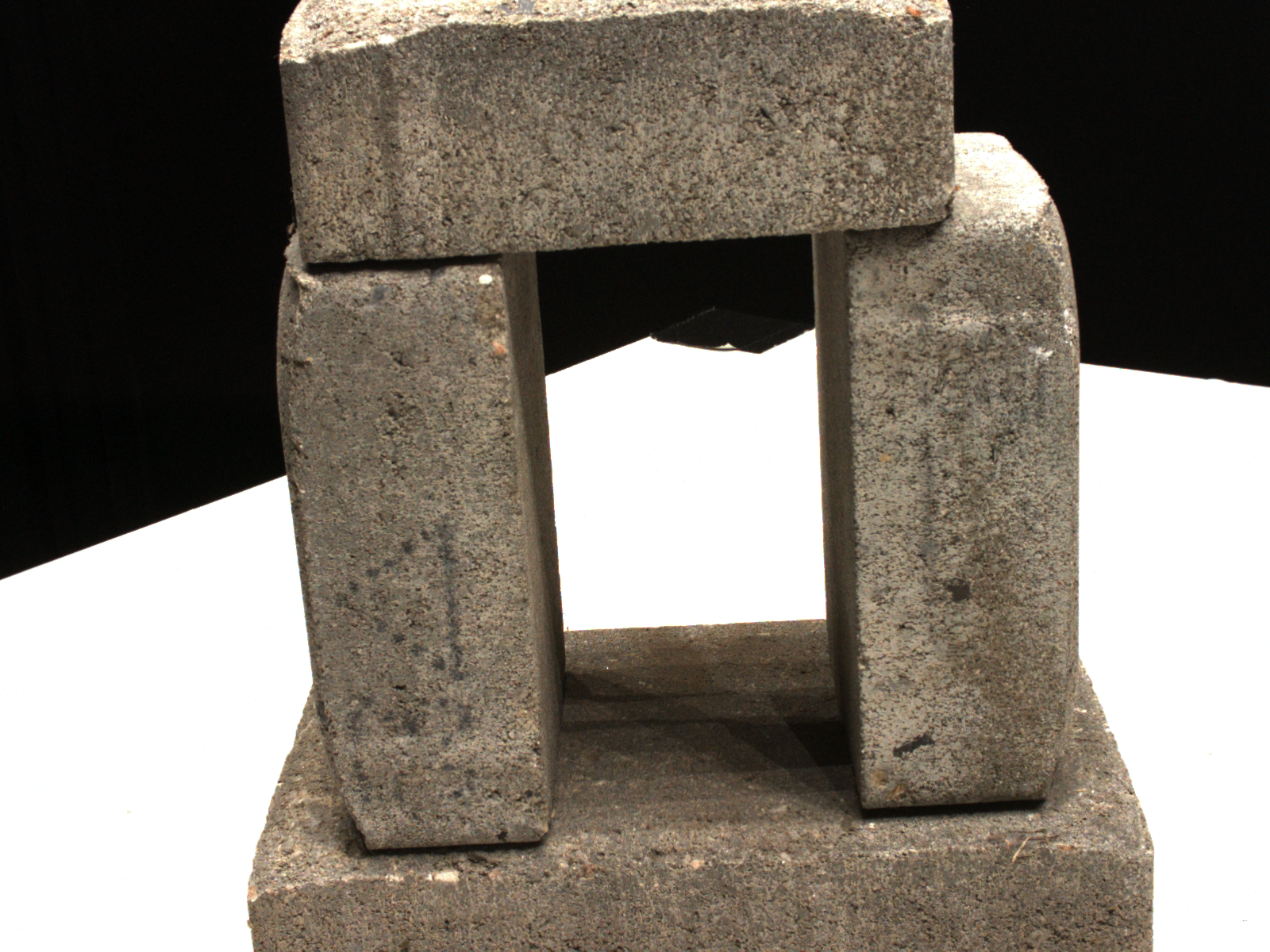}%
			\put(33, 35){\color{red}\bm{$\Box$}}
                \put(55, 35){\color{yellow}\bm{$\Box$}}%
		\end{overpic}\\[-0.5mm]%
		\emph{DTU40}
	}%
	\hspace{-1mm}\hfill%
	\mpage{0.8}{%
		\mpage{0.32}{%
			\includegraphics[width=\linewidth]{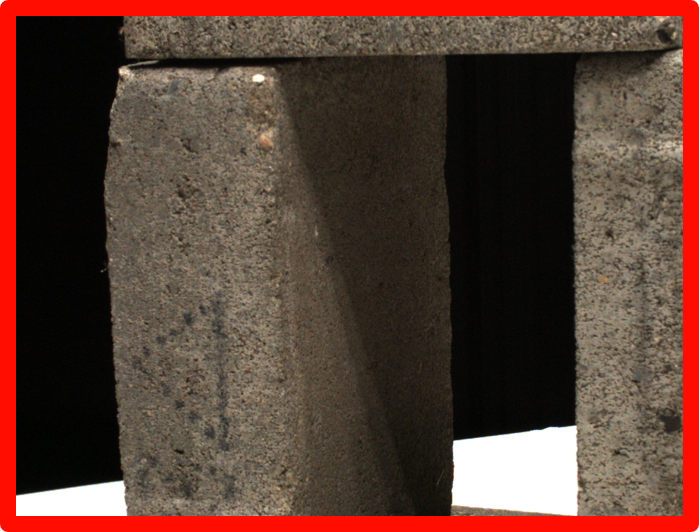}\\[0.2mm]%
		}\hspace{-1mm}\hfill%
		\mpage{0.32}{%
			\includegraphics[width=\linewidth]{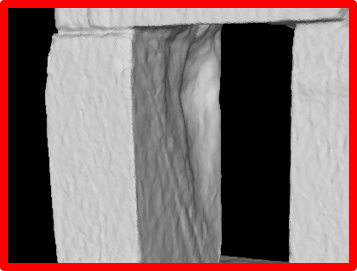}\\[0.2mm]%
		}\hspace{-1mm}\hfill%
		\mpage{0.32}{%
			\includegraphics[width=\linewidth]{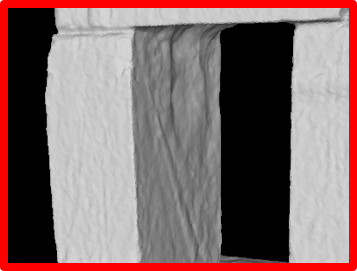}\\[0.2mm]%
		}
	}\\[1mm]%
         \mpage{0.19}{%
	}%
	\hspace{-1mm}\hfill%
        \mpage{0.8}{%
		\mpage{0.32}{%
			\includegraphics[width=\linewidth]{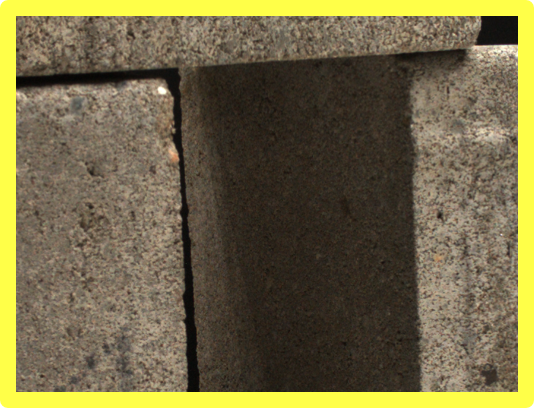}\\[0.2mm]%
		}\hspace{-1mm}\hfill%
		\mpage{0.32}{%
			\includegraphics[width=\linewidth]{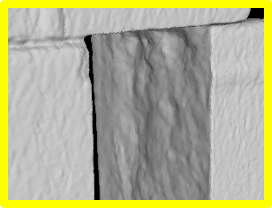}\\[0.2mm]%
		}\hspace{-1mm}\hfill%
		\mpage{0.32}{%
			\includegraphics[width=\linewidth]{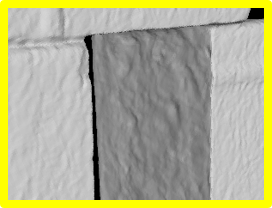}\\[0.2mm]%
		}
	}\\[1mm]%
	\mpage{0.19}{%
	}%
	\hspace{-1mm}\hfill%
	\mpage{0.8}{%
		\mpage{0.32}{%
			Reference%
		}\hspace{-1mm}\hfill%
		\mpage{0.32}{%
			NeuS~\cite{neus}%
		}\hspace{-1mm}\hfill%
		\mpage{0.32}{%
			w/ $L_0$-Sampler%
		}
	}\\[2mm]%
\caption{\textbf{Geometry Extraction Qualitative Comparison.} Our sampling refines NeuS~\cite{neus} to capture finer details and alleviate shading effects, improving geometry reconstruction accuracy.}
\label{fig:geo}
\end{figure}

Additionally, the quasi-$L_0$ $w(t)$ enables more focused sampling, resulting in a more precise capture of geometric details. That makes it beneficial for methods like NeuS~\cite{neus} that utilize importance sampling. The application of our $L_0$-Sampler with NeuS shows remarkable improvements in geometry, as depicted in \cref{fig:geo}, including correcting unnatural shading-induced pits (DTU 24 and 40) and capturing more challenging geometric details (DTU24 and \cref{fig:teaser}).

\noindent{\bf{Training Time Comparison. }}
As previously mentioned, our sampler has closed-form solutions at each step. In fact, it increases training time by less than 1\% compared to the original methods, thus delivering consistent improvements for negligible additional computational cost.

\subsection{Ablation Study on Interpolation Functions}
In our study, we evaluate several interpolation functions alongside those we initially proposed in \cref{oursolution}, with the results detailed in \cref{tab:splines}. Among these, the piecewise linear function ($\hat{w}(t) = a + (b-a)t$) was simple yet effective, although it does not focus sampling as intensely at the interval endpoints as our method. The piecewise exponential and inverse functions, both part of our proposal, show excellent performance in specific cases, particularly the exponential function for its consistent stability.
\renewcommand{\arraystretch}{1.}
\begin{table}[t]
    \begin{center}
    \resizebox{\linewidth}{!}{
    \begin{tabular}{|l|ccc|ccc|}
    \hline
                 \multirow{2}{*}{\textbf{Base Functions} } & \multicolumn{3}{c|}{\textbf{Chair}} & \multicolumn{3}{c|}{\textbf{Fern}} \\
    % \cline{2-7}
           & PSNR $\uparrow$ &SSIM $\uparrow$ & LPIPS $\downarrow$  & PSNR $\uparrow$ &SSIM $\uparrow$ & LPIPS $\downarrow$  \\
    \hline
    Constant &32.07&0.969&0.1050&26.33&0.857&0.1576\\
    Linear       &32.86&\underline{0.976}&\underline{0.0866}&26.41&0.859&0.1559\\
    Cubic Spline         &\underline{33.01}&0.976&0.0936&26.36&0.858&0.1566\\
    Akima           &32.96&0.975&0.0868&26.43&0.861&0.1518\\
    Inverse          &32.54&0.975&0.0956&\underline{26.54}&\underline{0.864}&\textbf{0.1472}\\
    Exponential          &\textbf{33.05}&\textbf{0.977}&\textbf{0.0852}&\textbf{26.56}&\textbf{0.864}&\underline{0.1473}\\
    \hline
    \end{tabular}}
    \end{center}
    \vspace{-5mm}
    \caption{{\bf Comparison between Interpolation Functions.} 
    We use Instant NGP training across different datasets, only changing the interpolation functions. The results indicate that the piecewise exponential function delivers the most consistent outcomes, while other functions also enhance the original HVS performance.
    }
    \label{tab:splines}
    \vspace{-3mm}
\end{table}

Additionally, we evaluate the cubic spline and Akima interpolation~\cite{akima}. Despite the fact that they are widely used, they do not offer the same level of performance, likely due to the specific requirements of $L_0$ property in real-world density function fitting. Furthermore, the computation of their polynomial coefficients results in increased computational complexity without producing significantly improved outcomes when compared to our $L_0$-Sampler. The choice of function may vary in practice, but our results indicate the potential of our proposed solutions.

\section{Conclusion and Future Work}

We have proposed $L_0$-Sampler and applied it to augment the hierarchical volume sampling (HVS), which is the most commonly used sampling strategy in NeRF. Different from previous studies, we adopt a piecewise exponential function to interpolate the weight function $w(t)$ during the sampling process and comprehensively evaluate the effectiveness of this approximation strategy. Its implementation only requires only a few lines of code modifications but can produce stable improvements to the NeRF series of works, which has been verified through extensive experiments.

\noindent{\bf{Limitations and Future Work}} Although our proposed $L_0$-Sampler improves performance in the vast majority of cases, not all results are improved. Currently, our method mainly improves the sampling strategy in the fine stage. The idea of how to apply the $L_0$ model in the coarse stage is also a research direction worth exploring.

{
    \small
    \bibliographystyle{ieeenat_fullname}
    \bibliography{main}

\begin{thebibliography}{65}
\providecommand{\natexlab}[1]{#1}
\providecommand{\url}[1]{\texttt{#1}}
\expandafter\ifx\csname urlstyle\endcsname\relax
  \providecommand{\doi}[1]{doi: #1}\else
  \providecommand{\doi}{doi: \begingroup \urlstyle{rm}\Url}\fi

\bibitem[Akima(1970)]{akima}
Hiroshi Akima.
\newblock A new method of interpolation and smooth curve fitting based on local procedures.
\newblock \emph{J. ACM}, 17:\penalty0 589--602, 1970.

\bibitem[Attal et~al.(2022)Attal, Huang, Zollh{\"o}fer, Kopf, and Kim]{lightfield}
Benjamin Attal, Jia-Bin Huang, Michael Zollh{\"o}fer, Johannes Kopf, and Changil Kim.
\newblock Learning neural light fields with ray-space embedding networks.
\newblock In \emph{Proceedings of the IEEE/CVF Conference on Computer Vision and Pattern Recognition (CVPR)}, 2022.

\bibitem[Barron et~al.(2021)Barron, Mildenhall, Tancik, Hedman, Martin-Brualla1, and Srinivasan]{mipnerf}
Jonathan~T. Barron, Ben Mildenhall, Matthew Tancik, Peter Hedman, Ricardo Martin-Brualla1, and Pratul~P. Srinivasan.
\newblock Mip-nerf: A multiscale representation for anti-aliasing neural radiance fields.
\newblock \emph{ICCV}, 2021.

\bibitem[Barron et~al.(2022)Barron, Mildenhall, Verbin, Srinivasan, and Hedman]{mip360}
Jonathan~T Barron, Ben Mildenhall, Dor Verbin, Pratul~P Srinivasan, and Peter Hedman.
\newblock Mip-nerf 360: Unbounded anti-aliased neural radiance fields.
\newblock \emph{Proceedings of the IEEE/CVF Conference on Computer Vision and Pattern Recognition}, page 5470–5479, 2022.

\bibitem[c and Zisserman(2021)]{nerf_in_detail}
Relja~Arandjelovi c and Andrew Zisserman.
\newblock Nerf in detail: Learning to sample for view synthesis.
\newblock \emph{arXiv preprint arXiv:2106.05264}, 2021.

\bibitem[Chen et~al.(2022)Chen, Xu, Geiger, Yu, and Su]{tensorf}
Anpei Chen, Zexiang Xu, Andreas Geiger, Jingyi Yu, and Hao Su.
\newblock Tensorf: Tensorial radiance fields.
\newblock In \emph{European Conference on Computer Vision (ECCV)}, 2022.

\bibitem[Dadon et~al.(2023)Dadon, Fried, and Hel-Or]{ddnerf}
David Dadon, Ohad Fried, and Yacov Hel-Or.
\newblock Ddnerf: Depth distribution neural radiance fields.
\newblock \emph{WACV}, 2023.

\bibitem[Du et~al.(2021)Du, Zhang, Yu, Tenenbaum, and Wu]{du2021nerflow}
Yilun Du, Yinan Zhang, Hong-Xing Yu, Joshua~B. Tenenbaum, and Jiajun Wu.
\newblock Neural radiance flow for 4d view synthesis and video processing.
\newblock In \emph{Proceedings of the IEEE/CVF International Conference on Computer Vision}, 2021.

\bibitem[Fang et~al.(2021)Fang, Xie, Wang, Zhang, Liu, and Tian]{neusample}
Jiemin Fang, Lingxi Xie, Xinggang Wang, Xiaopeng Zhang, Wenyu Liu, and Qi Tian.
\newblock Neusample: Neural sample field for efficient view synthesis.
\newblock \emph{arXiv preprint arXiv:2111.15552}, 2021.

\bibitem[Gao et~al.(2022)Gao, Zhong, Xiang, Hong, Guo, and Zhang]{nerfblendshape}
Xuan Gao, Chenglai Zhong, Jun Xiang, Yang Hong, Yudong Guo, and Juyong Zhang.
\newblock Reconstructing personalized semantic facial nerf models from monocular video.
\newblock \emph{ACM Transactions on Graphics (Proceedings of SIGGRAPH Asia)}, 41\penalty0 (6), 2022.

\bibitem[Gropp et~al.(2020)Gropp, Yariv, Haim, Atzmon, and Lipman]{igr}
Amos Gropp, Lior Yariv, Niv Haim, Matan Atzmon, and Yaron Lipman.
\newblock Implicit geometric regularization for learning shapes.
\newblock In \emph{Proceedings of Machine Learning and Systems 2020}, pages 3569--3579. 2020.

\bibitem[Guo et~al.(2021)Guo, Chen, Liang, Liu, Bao, and Zhang]{guo2021adnerf}
Yudong Guo, Keyu Chen, Sen Liang, Yongjin Liu, Hujun Bao, and Juyong Zhang.
\newblock Ad-nerf: Audio driven neural radiance fields for talking head synthesis.
\newblock In \emph{{IEEE/CVF} International Conference on Computer Vision (ICCV)}, 2021.

\bibitem[Hong et~al.(2022)Hong, Peng, Xiao, Liu, and Zhang]{headnerf}
Yang Hong, Bo Peng, Haiyao Xiao, Ligang Liu, and Juyong Zhang.
\newblock Headnerf: A real-time nerf-based parametric head model.
\newblock In \emph{{IEEE/CVF} Conference on Computer Vision and Pattern Recognition (CVPR)}. CVPR, 2022.

\bibitem[Jensen et~al.(2014)Jensen, Dahl, Vogiatzis, Tola, and Aanæs]{dtus}
Rasmus Jensen, Anders Dahl, George Vogiatzis, Engil Tola, and Henrik Aanæs.
\newblock Large scale multi-view stereopsis evaluation.
\newblock pages 406--413, 2014.

\bibitem[Jeong et~al.(2022)Jeong, Shin, and Park]{nerffactory}
Yoonwoo Jeong, Seungjoo Shin, and Kibaek Park.
\newblock nerf-factory: An awesome pytorch nerf library.
\newblock \emph{\url{https://github.com/kakaobrain/nerf-factory}}, 2022.

\bibitem[Kajiya and Von~Herzen(1984)]{raytracing}
James~T. Kajiya and Brian~P Von~Herzen.
\newblock Ray tracing volume densities.
\newblock 1984.

\bibitem[Karnewar et~al.(2022)Karnewar, Ritschel, Wang, and Mitra]{relufield}
Animesh Karnewar, Tobias Ritschel, Oliver Wang, and Niloy Mitra.
\newblock Relu fields: The little non-linearity that could.
\newblock In \emph{ACM SIGGRAPH 2022 Conference Proceedings}, New York, NY, USA, 2022. Association for Computing Machinery.

\bibitem[Kurz et~al.(2022)Kurz, Neff, Lv, Zollh\"{o}fer, and Steinberger]{adanerf}
Andreas Kurz, Thomas Neff, Zhaoyang Lv, Michael Zollh\"{o}fer, and Markus Steinberger.
\newblock Adanerf: Adaptive sampling for real-time rendering of neural radiance fields.
\newblock 2022.

\bibitem[Levoy(1990)]{efficient_sample}
M. Levoy.
\newblock Efficient ray tracing of volume data.
\newblock \emph{ACM Transactions on Graphics}, 1990.

\bibitem[Li et~al.(2023)Li, Gao, Tancik, and Kanazawa]{nerfacc}
Ruilong Li, Hang Gao, Matthew Tancik, and Angjoo Kanazawa.
\newblock Nerfacc: Efficient sampling accelerates nerfs.
\newblock \emph{arXiv preprint arXiv:2305.04966}, 2023.

\bibitem[Lindell et~al.(2021)Lindell, Martel, and Wetzstein]{autoint}
David~B. Lindell, Julien N.~P. Martel, and Gordon Wetzstein.
\newblock Autoint: Automatic integration for fast neural volume rendering.
\newblock 2021.

\bibitem[Lombardi et~al.(2019)Lombardi, Simon, Saragih, Schwartz, Lehrmann, and Sheikh]{nv}
Stephen Lombardi, Tomas Simon, Jason Saragih, Gabriel Schwartz, Andreas Lehrmann, and Yaser Sheikh.
\newblock Neural volumes: Learning dynamic renderable volumes from images.
\newblock \emph{ACM Trans. Graph.}, 38\penalty0 (4):\penalty0 65:1--65:14, 2019.

\bibitem[Lorensen and Cline(1987)]{mcube}
William~E. Lorensen and Harvey~E. Cline.
\newblock Marching cubes: A high resolution 3d surface construction algorithm.
\newblock \emph{SIGGRAPH Comput. Graph.}, 21\penalty0 (4):\penalty0 163–169, 1987.

\bibitem[Marschner and Lobb(1994)]{volume_filter}
S.R. Marschner and R.J. Lobb.
\newblock An evaluation of reconstruction filters for volume rendering.
\newblock In \emph{Proceedings Visualization '94}, pages 100--107, 1994.

\bibitem[Mildenhall et~al.(2019)Mildenhall, Srinivasan, Ortiz-Cayon, Kalantari, Ramamoorthi, Ng, and Kar]{llff}
Ben Mildenhall, Pratul~P. Srinivasan, Rodrigo Ortiz-Cayon, Nima~Khademi Kalantari, Ravi Ramamoorthi, Ren Ng, and Abhishek Kar.
\newblock Local light field fusion: Practical view synthesis with prescriptive sampling guidelines.
\newblock \emph{ACM Transactions on Graphics (TOG)}, 2019.

\bibitem[Mildenhall et~al.(2020)Mildenhall, Srinivasan, Tancik, Barron, Ramamoorthi, and Ng]{nerf}
Ben Mildenhall, Pratul~P. Srinivasan, Matthew Tancik, Jonathan~T Barron, Ravi Ramamoorthi, and Ren Ng.
\newblock Nerf: Representing scenes as neural radiance fields for view synthesis.
\newblock \emph{ECCV}, 2020.

\bibitem[Morozov et~al.(2023)Morozov, Rakitin, Desheulin, Vetrov, and Struminsky]{reparameterized}
Nikita Morozov, Denis Rakitin, Oleg Desheulin, Dmitry Vetrov, and Kirill Struminsky.
\newblock Differentiable rendering with reparameterized volume sampling.
\newblock \emph{arXiv preprint arXiv:2302.10970}, 2023.

\bibitem[M\"uller et~al.(2022)M\"uller, Evans, Schied, and Keller]{instantngp}
Thomas M\"uller, Alex Evans, Christoph Schied, and Alexander Keller.
\newblock Instant neural graphics primitives with a multiresolution hash encoding.
\newblock \emph{ACM Trans. Graph.}, 41\penalty0 (4):\penalty0 102:1--102:15, 2022.

\bibitem[Neff et~al.(2021)Neff, Stadlbauer, Parger, Kurz, Mueller, Chaitanya, Kaplanyan, and Steinberger]{donerf}
Thomas Neff, Pascal Stadlbauer, Mathias Parger, Andreas Kurz, Joerg~H. Mueller, Chakravarty R.~Alla Chaitanya, Anton~S. Kaplanyan, and Markus Steinberger.
\newblock Donerf: Towards real-time rendering of compact neural radiance fields using depth oracle networks.
\newblock \emph{Computer Graphics Forum}, 2021.

\bibitem[Niemeyer et~al.(2020)Niemeyer, Mescheder, Oechsle, and Geiger]{DVR}
Michael Niemeyer, Lars Mescheder, Michael Oechsle, and Andreas Geiger.
\newblock Differentiable volumetric rendering: Learning implicit 3d representations without 3d supervision.
\newblock In \emph{Proc. IEEE Conf. on Computer Vision and Pattern Recognition (CVPR)}, 2020.

\bibitem[Noguchi et~al.(2021)Noguchi, Sun, Lin, and Harada]{narf}
Atsuhiro Noguchi, Xiao Sun, Stephen Lin, and Tatsuya Harada.
\newblock Neural articulated radiance field.
\newblock In \emph{International Conference on Computer Vision}, 2021.

\bibitem[Oechsle et~al.(2021)Oechsle, Peng, and Geiger]{unisurf}
Michael Oechsle, Songyou Peng, and Andreas Geiger.
\newblock Unisurf: Unifying neural implicit surfaces and radiance fields for multi-view reconstruction.
\newblock In \emph{International Conference on Computer Vision (ICCV)}, 2021.

\bibitem[Park et~al.(2019)Park, Florence, Straub, Newcombe, and Lovegrove]{deepsdf}
Jeong~Joon Park, Peter Florence, Julian Straub, Richard Newcombe, and Steven Lovegrove.
\newblock Deepsdf: Learning continuous signed distance functions for shape representation.
\newblock In \emph{The IEEE Conference on Computer Vision and Pattern Recognition (CVPR)}, 2019.

\bibitem[Park et~al.(2021{\natexlab{a}})Park, Sinha, Barron, Bouaziz, Goldman, Seitz, and Martin-Brualla]{nerfies}
Keunhong Park, Utkarsh Sinha, Jonathan~T. Barron, Sofien Bouaziz, Dan~B Goldman, Steven~M. Seitz, and Ricardo Martin-Brualla.
\newblock Nerfies: Deformable neural radiance fields.
\newblock \emph{ICCV}, 2021{\natexlab{a}}.

\bibitem[Park et~al.(2021{\natexlab{b}})Park, Sinha, Hedman, Barron, Bouaziz, Goldman, Martin-Brualla, and Seitz]{hypernerf}
Keunhong Park, Utkarsh Sinha, Peter Hedman, Jonathan~T. Barron, Sofien Bouaziz, Dan~B Goldman, Ricardo Martin-Brualla, and Steven~M. Seitz.
\newblock Hypernerf: A higher-dimensional representation for topologically varying neural radiance fields.
\newblock \emph{ACM Trans. Graph.}, 40\penalty0 (6), 2021{\natexlab{b}}.

\bibitem[Peng et~al.(2023)Peng, Hu, Zhou, Gao, and Zhang]{intrinsicngp}
Bo Peng, Jun Hu, Jingtao Zhou, Xuan Gao, and Juyong Zhang.
\newblock Intrinsicngp: Intrinsic coordinate based hash encoding for human nerf.
\newblock \emph{IEEE Transactions on Visualization and Computer Graphics}, 2023.

\bibitem[Piala and Clark(2021)]{terminerf}
M. Piala and R. Clark.
\newblock Terminerf: Ray termination prediction for efficient neural rendering.
\newblock pages 1106--1114, 2021.

\bibitem[Pumarola et~al.(2021)Pumarola, Corona, Pons-Moll, and Moreno-Noguer]{dnerf}
Albert Pumarola, Enric Corona, Gerard Pons-Moll, and Francesc Moreno-Noguer.
\newblock D-nerf: Neural radiance fields for dynamic scenes.
\newblock In \emph{Proceedings of the IEEE/CVF Conference on Computer Vision and Pattern Recognition}, 2021.

\bibitem[{Sara Fridovich-Keil and Alex Yu} et~al.(2022){Sara Fridovich-Keil and Alex Yu}, Tancik, Chen, Recht, and Kanazawa]{Plenoxels}
{Sara Fridovich-Keil and Alex Yu}, Matthew Tancik, Qinhong Chen, Benjamin Recht, and Angjoo Kanazawa.
\newblock Plenoxels: Radiance fields without neural networks.
\newblock In \emph{CVPR}, 2022.

\bibitem[Sitzmann et~al.(2021)Sitzmann, Rezchikov, Freeman, Tenenbaum, and Durand]{lightfieldnet}
Vincent Sitzmann, Semon Rezchikov, William~T. Freeman, Joshua~B. Tenenbaum, and Fredo Durand.
\newblock Light field networks: Neural scene representations with single-evaluation rendering.
\newblock In \emph{Proc. NeurIPS}, 2021.

\bibitem[Srinivasan et~al.(2021)Srinivasan, Deng, Zhang, Tancik, Mildenhall, and Barron]{nerfrv}
P.~P. Srinivasan, B. Deng, X. Zhang, M. Tancik, B. Mildenhall, and J.~T. Barron.
\newblock Nerv: Neural reflectance and visibility fields for relighting and view synthesis.
\newblock pages 7491--7500, 2021.

\bibitem[Sun et~al.(2022)Sun, Sun, and Chen]{dvgo}
Cheng Sun, Min Sun, and Hwann{-}Tzong Chen.
\newblock Direct voxel grid optimization: Super-fast convergence for radiance fields reconstruction.
\newblock \emph{CVPR}, 2022.

\bibitem[Sun et~al.(2023)Sun, Liu, Fan, Liu, Dong, and Kong]{efficient_rays}
Shilei Sun, Ming Liu, Zhongyi Fan, Yuxue Liu, Liquan Dong, and Lingqin Kong.
\newblock Efficient ray sampling for radiance fields reconstruction.
\newblock \emph{arXiv:2308.15547}, 2023.

\bibitem[Takikawa et~al.(2022)Takikawa, Perel, Tsang, Loop, Litalien, Tremblay, Fidler, and Shugrina]{kaolin}
Towaki Takikawa, Or Perel, Clement~Fuji Tsang, Charles Loop, Joey Litalien, Jonathan Tremblay, Sanja Fidler, and Maria Shugrina.
\newblock Kaolin wisp: A pytorch library and engine for neural fields research.
\newblock \emph{\url{https://github.com/NVIDIAGameWorks/kaolin-wisp}}, 2022.

\bibitem[Tancik et~al.(2023)Tancik, Weber, Ng, Li, Yi, Kerr, Wang, Kristoffersen, Austin, Salahi, Ahuja, McAllister, and Kanazawa]{nerfstudio}
Matthew Tancik, Ethan Weber, Evonne Ng, Ruilong Li, Brent Yi, Justin Kerr, Terrance Wang, Alexander Kristoffersen, Jake Austin, Kamyar Salahi, Abhik Ahuja, David McAllister, and Angjoo Kanazawa.
\newblock Nerfstudio: A modular framework for neural radiance field development.
\newblock \emph{ACM SIGGRAPH 2023 Conference Proceedings}, 2023.

\bibitem[Tang(2022)]{torchngp}
Jiaxiang Tang.
\newblock Torch-ngp: a pytorch implementation of instant-ngp, 2022.
\newblock https://github.com/ashawkey/torch-ngp.

\bibitem[Tang et~al.(2022)Tang, Chen, Wang, and Zeng]{tang2022compressible}
Jiaxiang Tang, Xiaokang Chen, Jingbo Wang, and Gang Zeng.
\newblock Compressible-composable nerf via rank-residual decomposition.
\newblock \emph{arXiv preprint arXiv:2205.14870}, 2022.

\bibitem[Tong et~al.(2023)Tong, Muthu, Maken, Nguyen, and Li]{glass}
Jinguang Tong, Sundaram Muthu, Fahira~Afzal Maken, Chuong Nguyen, and Hongdong Li.
\newblock Seeing through the glass: Neural 3d reconstruction of object inside a transparent container.
\newblock In \emph{Proceedings of the IEEE/CVF Conference on Computer Vision and Pattern Recognition (CVPR)}, pages 12555--12564, 2023.

\bibitem[Tretschk et~al.(2020)Tretschk, Tewari, Golyanik, Zollhöfer, Lassner, and Theobalt]{nonrigid}
Edgar Tretschk, Ayush Tewari, Vladislav Golyanik, Michael Zollhöfer, Christoph Lassner, and Christian Theobalt.
\newblock Non-rigid neural radiance fields: Reconstruction and novel view synthesis of a dynamic scene from monocular video.
\newblock 2020.

\bibitem[Ueda et~al.(2022)Ueda, Fukuhara, Kataoka, Aizawa, Shishido, and Kitahara]{nddf}
Itsuki Ueda, Yoshihiro Fukuhara, Hirokatsu Kataoka, Hiroaki Aizawa, Hidehiko Shishido, and Itaru Kitahara.
\newblock Neural density-distance fields.
\newblock In \emph{Proceedings of the European Conference on Computer Vision}, 2022.

\bibitem[Wang et~al.(2021)Wang, Liu, Liu, Theobalt, Komura, and Wang]{neus}
Peng Wang, Lingjie Liu, Yuan Liu, Christian Theobalt, Taku Komura, and Wenping Wang.
\newblock Neus: Learning neural implicit surfaces by volume rendering for multi-view reconstruction.
\newblock \emph{NeurIPS}, 2021.

\bibitem[Wang et~al.(2022)Wang, Liu, Lin, Gu, Liu, Komura, and Wang]{lumigraph}
Peng Wang, Yuan Liu, Guying Lin, Jiatao Gu, Lingjie Liu, Taku Komura, and Wenping Wang.
\newblock Progressively-connected light field network for efficient view synthesis.
\newblock \emph{arXiv preprint arXiv:2207.04465}, 2022.

\bibitem[Wang et~al.(2004)Wang, Bovik, Sheikh, and Simoncelli]{ssim}
Zhou Wang, A.C. Bovik, H.R. Sheikh, and E.P. Simoncelli.
\newblock Image quality assessment: from error visibility to structural similarity.
\newblock \emph{IEEE Transactions on Image Processing}, 13\penalty0 (4):\penalty0 600--612, 2004.

\bibitem[Xu et~al.(2022)Xu, Xu, Philip, Bi, Shu, Sunkavalli, and Neumann]{pointnerf}
Qiangeng Xu, Zexiang Xu, Julien Philip, Sai Bi, Zhixin Shu, Kalyan Sunkavalli, and Ulrich Neumann.
\newblock Point-nerf: Point-based neural radiance fields.
\newblock In \emph{Proceedings of the IEEE/CVF Conference on Computer Vision and Pattern Recognition}, pages 5438--5448, 2022.

\bibitem[Yao et~al.(2020)Yao, Luo, Li, Zhang, Ren, Zhou, Fang, and Quan]{bmvs}
Yao Yao, Zixin Luo, Shiwei Li, Jingyang Zhang, Yufan Ren, Lei Zhou, Tian Fang, and Long Quan.
\newblock Blendedmvs: A large-scale dataset for generalized multi-view stereo networks.
\newblock pages 1787--1796, 2020.

\bibitem[Yariv et~al.(2020)Yariv, Kasten, Moran, Galun, Atzmon, Ronen, and Lipman]{idr}
Lior Yariv, Yoni Kasten, Dror Moran, Meirav Galun, Matan Atzmon, Basri Ronen, and Yaron Lipman.
\newblock Multiview neural surface reconstruction by disentangling geometry and appearance.
\newblock \emph{Advances in Neural Information Processing Systems}, 33, 2020.

\bibitem[Yariv et~al.(2021)Yariv, Gu, Kasten, and Lipman]{volumesdf}
Lior Yariv, Jiatao Gu, Yoni Kasten, and Yaron Lipman.
\newblock Volume rendering of neural implicit surfaces.
\newblock In \emph{Thirty-Fifth Conference on Neural Information Processing Systems}, 2021.

\bibitem[Yu et~al.(2021{\natexlab{a}})Yu, Li, Tancik, Li, Ng, and Kanazawa]{plenoctrees}
Alex Yu, Ruilong Li, Matthew Tancik, Hao Li, Ren Ng, and Angjoo Kanazawa.
\newblock {PlenOctrees} for real-time rendering of neural radiance fields.
\newblock In \emph{ICCV}, 2021{\natexlab{a}}.

\bibitem[Yu et~al.(2021{\natexlab{b}})Yu, Ye, Tancik, and Kanazawa]{pixelnerf}
Alex Yu, Vickie Ye, Matthew Tancik, and Angjoo Kanazawa.
\newblock pixelnerf: Neural radiance fields from one or few images.
\newblock In \emph{CVPR}, 2021{\natexlab{b}}.

\bibitem[Y{\"{u}}cer et~al.(2016)Y{\"{u}}cer, Sorkine{-}Hornung, Wang, and Sorkine{-}Hornung]{lfdata}
Kaan Y{\"{u}}cer, Alexander Sorkine{-}Hornung, Oliver Wang, and Olga Sorkine{-}Hornung.
\newblock Efficient 3d object segmentation from densely sampled light fields with applications to 3d reconstruction.
\newblock \emph{{ACM} Trans. Graph.}, 35\penalty0 (3):\penalty0 22:1--22:15, 2016.

\bibitem[Zhang et~al.(2020)Zhang, Riegler, Snavely, and Koltun]{nerfpp}
Kai Zhang, Gernot Riegler, Noah Snavely, and Vladlen Koltun.
\newblock Nerf++: Analyzing and improving neural radiance fields.
\newblock \emph{arXiv:2010.07492}, 2020.

\bibitem[Zhang(2019)]{maxblur}
Richard Zhang.
\newblock Making convolutional networks shift-invariant again.
\newblock \emph{ICML}, 2019.

\bibitem[Zhang et~al.(2018)Zhang, Isola, Efros, Shechtman, and Wang]{lpips}
Richard Zhang, Phillip Isola, Alexei Efros, Eli Shechtman, and Oliver Wang.
\newblock The unreasonable effectiveness of deep features as a perceptual metric.
\newblock 2018.

\bibitem[Zhang et~al.(2023)Zhang, Xing, Zeng, Liu, Shi, and Han]{few_rays}
Wenyuan Zhang, Ruofan Xing, Yunfan Zeng, Yu-Shen Liu, Kanle Shi, and Zhizhong Han.
\newblock Fast learning radiance fields by shooting much fewer rays.
\newblock \emph{IEEE Transactions on Image Processing}, 2023.

\bibitem[Zhang et~al.(2021)Zhang, Srinivasan, Deng, Debevec, Freeman, and Barron]{NeRFactor}
Xiuming Zhang, Pratul~P. Srinivasan, Boyang Deng, Paul Debevec, William~T. Freeman, and Jonathan~T. Barron.
\newblock Nerfactor: Neural factorization of shape and reflectance under an unknown illumination.
\newblock \emph{TOG 2021 (Proc. SIGGRAPH Asia)}, 2021.

\end{thebibliography}
}

\end{document}